\documentclass[12pt]{article}

\usepackage[colorlinks=true,
            linkcolor=red,
            urlcolor=blue,
            citecolor=blue,
            breaklinks=true]{hyperref}
\usepackage{breakurl} 
\usepackage[a4paper, left=3cm, right=3cm, top=3cm, bottom=3cm]{geometry}
\usepackage{titlesec}
\usepackage{amssymb}
\usepackage{setspace}
\usepackage{dsfont}
\usepackage{natbib}
\usepackage{amsmath}
\usepackage{graphicx}
\usepackage{booktabs,multirow,array}
\usepackage{algorithm}
\usepackage{caption}
\usepackage{rotating}
\usepackage{subcaption}
\usepackage{url}
\DeclareCaptionFormat{myformat}{#3}
\usepackage{algpseudocode}
\newtheorem{theorem}{Theorem}

\usepackage[dvipsnames]{xcolor}

\newcommand{\dd}{\mathrm{d}}

\newcommand{\bP}{\mathbb P}
\newcommand{\E}{\mathbb E}
\newcommand{\N}{\mathbb N}
\newcommand{\C}{\mathbb{C}}
\newcommand{\D}{\mathbb{D}}
\newcommand{\R}{\mathbb R}
\newcommand{\cA}{\mathcal A}
\newcommand{\cB}{\mathcal B}
\newcommand{\cM}{\mathcal M}
\newcommand{\cC}{\mathcal C}
\newcommand{\cG}{\mathcal G}
\newcommand{\cH}{\mathcal H}
\newcommand{\cN}{\mathcal N}
\newcommand{\cU}{\mathcal U}
\newcommand{\Gb}{\bar G}
\newcommand{\Fb}{\bar F}

\newcommand{\by}{\boldsymbol y}
\newcommand{\bz}{\boldsymbol z}
\newcommand{\bdelta}{\boldsymbol \delta}
\newcommand{\bSigma}{\textbf{$\Sigma$}}
\newcommand{\norm}[1]{\|#1\|}

\newtheorem{hyp}{Hypothesis}

\date{}

\begin{document}

\title{\vspace{-1cm} \fontsize{19}{19}\selectfont C-mix: a high dimensional mixture model for censored durations, with applications to genetic data \vspace{-1.5cm}}

\maketitle

\begin{center}
{\setstretch{1.1}

\large Simon Bussy \\
\normalsize
Theoretical and Applied Statistics Laboratory\\
Pierre and Marie Curie University, Paris, France\\
\emph{email}: \texttt{simon.bussy@gmail.com}\\

\vspace{0.2cm}

\large Agathe Guilloux \\
\normalsize
LaMME, UEVE and UMR 8071\\ 
Paris Saclay University, Evry, France \\
\emph{email}: \texttt{agathe.guilloux@math.cnrs.fr}\\

\vspace{0.2cm}

\large St\'ephane Ga\"iffas \\
\normalsize
CMAP Ecole polytechnique, Palaiseau, France \\ 
and LPMA, UMR CNRS 7599, Paris Diderot University, Paris, France \\
\emph{email}: \texttt{stephane.gaiffas@polytechnique.edu}\\

\vspace{0.2cm}

\large Anne-Sophie Jannot \\
\normalsize
Biomedical Informatics and Public Health Department \\  European Georges Pompidou Hospital, Assistance Publique-H\^opitaux de Paris \\
and INSERM UMRS 1138, Centre de Recherche des Cordeliers, Paris, France \\ 
\emph{email}: \texttt{annesophie.jannot@aphp.fr}\\
}
\vspace{.3cm}
\end{center}

\begin{abstract}
{We introduce a supervised learning mixture model for censored durations (C-mix) to simultaneously detect subgroups of patients with different prognosis and order them based on their risk. Our method is applicable in a high-dimensional setting, i.e. with a large number of biomedical covariates. 
Indeed, we penalize the negative log-likelihood by the Elastic-Net, which leads to a sparse parameterization of the model and automatically pinpoints the relevant covariates for the survival prediction.
Inference is achieved using an efficient Quasi-Newton Expectation Maximization (QNEM) algorithm, for which we provide convergence properties.
The statistical performance of the method is examined on an extensive Monte Carlo simulation study, and finally illustrated on three publicly available genetic cancer datasets with high-dimensional covariates. 
We show that our approach outperforms the state-of-the-art survival models in this context, namely both the CURE and Cox proportional hazards models penalized by the Elastic-Net, in terms of C-index, AUC($t$) and survival prediction.
Thus, we propose a powerfull tool for personalized medicine in cancerology.} \\

\noindent
\emph{Keywords.} Cox’s proportional hazards model; CURE model; Elastic-net regularization; High-dimensional estimation; Mixture duration model; Survival analysis
\end{abstract}

\section{Introduction}
\label{section intro}

Predicting subgroups of patients with different prognosis is a key challenge for personalized medicine, see for instance \citet{alizadeh2000distinct} and \citet{rosenwald2002use} where subgroups of patients with different survival rates are identified based on gene expression data.
A substantial number of techniques can be found in the literature to predict the subgroup of a given patient in a classification setting, namely when subgroups are known in advance \citep{golub1999molecular, hastie2001supervised, tibshirani2002diagnosis}. We consider in the present paper the much more difficult case where subgroups are unknown. 

In this situation, a first widespread approach consists in first using unsupervised learning techniques applied on the covariates -- for instance on the gene expression data \citep{bhattacharjee2001classification, beer2002gene, sorlie2001gene} --  to define subsets of patients and then estimating the risks in each of them.
The problem of such techniques is that there is no guarantee that the identified subgroups will have different risks.
Another approach to subgroups identification is conversely based exclusively on the survival times: patients are then assigned to a “low-risk” or a “high-risk” group based on whether they were still alive \citep{shipp2002diffuse, van2002gene}. The problem here is that the resulting subgroups may not be biologically meaningful since the method do not use the covariates, and prognosis prediction based on covariates is not possible.

The method we propose uses both the survival information of the patients and its covariates in a supervised learning way. Moreover, it relies on the idea that exploiting the subgroups structure of the data, namely the fact that a portion of the population have a higher risk of early death, could improve the survival prediction of future patients (unseen during the learning phase). 

We propose to consider a mixture of event times distributions in which the probabilities of belonging to each subgroups are driven by the covariates (e.g. gene expression data, patients characteristics, therapeutic strategy or omics covariates). Our C-mix model is hence part of the class of model-based clustering algorithms, as introduced in~\citet{banfield1993model}.

 More precisely, to model the heterogeneity within the patient population, we introduce a latent variable $Z \in \{ 0, \ldots, K-1\}$ and our focus is on the conditional distribution of $Z$ given the values of the covariates $X=x$. Now, conditionally on the latent variable $Z$, the distribution of duration time $T$ is different, leading to a mixture in the event times distribution. 

For a patient with covariates $x$, the conditional probabilities $\pi_k(x) = \bP[Z = k | X=x]$  of belonging to the $k$-th risk group can be seen as scores, that can help decision-making for physicians. As a byproduct, it can also shed light on the effect of the covariates (which combination of biomedical markers are relevant to a given event of interest).

Our methodology differs from the standard survival analysis approaches in various ways, that we describe in this paragraph.
First, the Cox proportional hazards (PH) model (\citet{Cox1972JRSS}) (by far the most widely used in such a setting) is a regression model that describes the relation between intensity of events and covariates $x$ via
\begin{equation}
\label{eq:cox-intensity}
  \lambda(t) = \lambda_0(t) \text{exp}(x^\top \beta^{\text{cox}}),
\end{equation}
where $\lambda_0$ is a baseline intensity, and $\beta^\text{cox}$ is a vector quantifying the multiplicative impact on the hazard ratio of each covariate. As in our model, high-dimensional covariates can be handled, via e.g. penalization , see~\citet{simon2011regularization}. But it does not permit the stratification of the population in groups of homogeneous risks, hence does no deliver a simple tool for clinical practice. Moreover, we show in the numerical sections that the C-mix model can be trained very efficiently in high dimension, and outperforms the standard Cox PH model by far in the analysed datasets.

Other models condiser mixtures of event times distributions. In the CURE model (see~\citet{farewell1982use} and \citet{kuk1992mixture}), one fraction of the population is considered as cured (hence not subject to any risk). This can be very limitating, as for a large number of applications (e.g. rehospitalization for patients with chronic diseases or relapse for patients with metastatic cancer), all patients are at risk. We consider, in our model, that there is always an event risk, no matter how small.
Other mixture models have been considered in survival analysis: see \citet{kuo2000mixture} for a general study about mixture model for survival data or \citet{de1999mixture} in a cancer survival analysis setting, to name but a few. Unlike our algorithm, none of these algorithms considers the high dimensional setting.

A precise description of the model is given in Section~\ref{sec:mix_model}. Section~\ref{sec:inference} focuses on the regularized version of the model with an Elastic-Net penalization to exploit dimension reduction and prevent overfitting. Inference is presented under this framework, as well as the convergence properties of the developed algorithm. Section~\ref{sec:Performance evaluation} highlights the simulation procedure used to evaluate the performances and compares it with state-of-the-art models. In Section~\ref{sec:application}, we apply our method to genetic datasets. Finally, we discuss the obtained results in Section~\ref{sec:concluding remarks}.

\section{A censored mixture model}
\label{sec:mix_model}
Let us present the survival analysis framework. We assume that, the conditional density of the duration $T$ given $X=x$ is a mixture
\begin{equation*}
  f(t|X=x) = \sum_{k=0}^{K-1} \pi_k(x) f_k(t ; \alpha_k)
\end{equation*}
of $K \geq 1$ densities $f_k$, for $t \geq 0$ and $\alpha_k \in \R^{d_k}$ some parameters to estimate.
The weights combining these distributions depend on the patient biomedical covariates $x$ and are such that 
\begin{equation}
\label{eq:sum-to-1}
\sum_{k=0}^{K-1} \pi_k(x) = 1.
\end{equation}This is equivalent to saying that conditionally on a latent variable $Z=k \in \{ 0, \ldots, K-1 \}$, the density of $T$ at time $t \geq 0$ is $f_k(t\ ; \alpha_k)$, and we have
\begin{equation*}
  \bP[Z = k | X=x] = \pi_k(x) = \pi_{\beta_k}(x)
\end{equation*}
where $\beta_k = (\beta_{k,1}, \ldots, \beta_{k,d}) \in \R^d$ denotes a vector of coefficients that quantifies the impact of each biomedical covariates on the probability that a patient belongs to the $k$-th group.
Consider a logistic link function for these weights given by
\begin{equation}
  \label{eq:pi}
  \pi_{\beta_k}(x) = \frac{e^{x^\top \beta_k}}{\sum_{k=0}^{K-1} e^{x^\top \beta_k}}.
\end{equation} 
The hidden status $Z$ has therefore a multinomial distribution $\cM\big(\pi_{\beta_0}(x), \ldots, \pi_{\beta_{K-1}}(x)\big)$.
The intercept term is here omitted without loss of generality.

In practice, information loss occurs of right censoring type. This is taken into acount in our model by introducing the following: a time $C \geq 0$ when the individual ``leaves'' the target cohort, a right-censored duration $Y$ and a censoring indicator $\Delta$, defined by
\begin{equation*}
  Y = \text{min}(T,C) \;\; \text{ and } \;\; \Delta = \mathds{1}_{\{T \leq C\}},
\end{equation*}
where $\text{min}(a,b)$ denotes the minimum between two numbers $a$ and $b$, and $\mathds{1}$ denotes the indicator function.

In order to write a likelihood and draw inference, we make the two following hypothesis.
\begin{hyp}\label{hyp:first}
$T$ and $C$ are conditionally independent given $Z$ and $X$.
\end{hyp}
\begin{hyp}\label{hyp:second}
$C$ is independent of $Z$.
\end{hyp}
Hypothesis~\ref{hyp:first} is classical in survival analysis \citep{klein2005survival}, while Hypothesis~\ref{hyp:second} is classical in survival mixture models \citep{kuo2000mixture, de1999mixture}.
Under this hypothesis, denoting $g$ the density of the censoring $C$, $F$ the cumulative distribution function corresponding to a given density $f$, $\Fb=1-F$ and $F(y^-) = \underset{\underset{u \leq y}{u \rightarrow y}}{\text{lim}} F(u)$, we have
\begin{align*}
\bP[Y \leq y, \Delta=1] &= \bP[T \leq y, T \leq C] = \int_0^y f(u)\Gb(u) \dd u  \quad \text{and} \\
\bP[Y \leq y, \Delta=0] &= \bP[C \leq y, C < T] = \int_0^y g(u)\Fb(u) \dd u.
\end{align*}
Then, denoting $\theta = (\alpha_0, \ldots, \alpha_{K-1}, \beta_0, \ldots, \beta_{K-1})^\top$ the parameters to infer and considering an independent and identically distributed (i.i.d.) cohort of $n$ patients $(x_1,y_1,\delta_1), \dots , (x_n,y_n,\delta_n) \in \R^d \times \R_+ \times \{0,1\}$, the log-likelihood of the C-mix model can be written
\begin{align*}
\ell_n(\theta) = \ell_n(\theta\ ; \by,\bdelta)  
= n^{-1} \sum_{i=1}^n \Big\{ \delta_i &\log \Big[ \Gb(y_i^-) \sum_{k=0}^{K-1} \pi_{\beta_k}(x_i) f_k(y_i;\alpha_k) \Big] \\
  \quad \quad \quad \quad \quad + (1-\delta_i) &\log \Big[ g(y_i) \sum_{k=0}^{K-1} \pi_{\beta_k}(x_i) \Fb_k(y_i^-;\alpha_k) \Big] \Big\},
\end{align*}
where we use the notations $\by = (y_1, \ldots , y_n)^\top$ and $\bdelta = (\delta_1, \dots ,\delta_n)^\top$. Note that from now on, all computations are done conditionally on the covariates $(x_i)_{i=1, \dots, n}$. An important fact is that we \emph{do not need to know or parametrize} $\Gb$ nor $g$, namely the distribution of the censoring, for inference in this model (since all $\Gb$ and $g$ terms vanish in Equation~\eqref{eq:qi}).
  
\section{Inference of C-mix \label{sec:inference}}
In this section, we describe the procedure for estimating the parameters of the C-mix model. We begin by presenting the Quasi-Newton Expectation Maximization (QNEM) algorithm we use for inference. We then focus our study on the convergence properties of the algorithm.

\subsection{QNEM algorithm \label{QNEM algorithm}}
In order to avoid overfitting and to improve the prediction power of our model, we use Elastic-Net regularization \citep{zou2005regularization} by minimizing the penalized objective
\begin{equation}
  \label{eq:objective}
  \ell^{\text{pen}}_n(\theta) = - \ell_n(\theta) + \sum_{k=0}^{K-1} \gamma_k \big( (1-\eta)\norm{\beta_k}_1 + \frac \eta 2 \norm{\beta_k}_2^2 \big),
\end{equation}
where we add a linear combination of the lasso ($\ell_1$) and ridge (squared $\ell_2$) penalties for a fixed $\eta \in [0,1]$, tuning parameter $\gamma_k$, and where we denote $\norm{\beta_k}_p = \big( \sum_{i=1}^d |\beta_{k,i}|^p \big)^{1 / p}$ the $\ell_p$-norm of $\beta_k$. One advantage of this regularization method is its ability to perform model selection (the lasso part) and pinpoint the most important covariates relatively to the prediction objective. On the other hand, the ridge part allows to handle potential correlation between covariates \citep{zou2005regularization}. Note that in practice, the intercept is not regularized.

In order to derive an algorithm for this objective, we introduce a so-called Quasi-Newton Expectation Maximization (QNEM), being a combination between an EM algorithm \citep{dempster1977maximum} and a L-BFGS-B algorithm \citep{zhu1997algorithm}. For the EM part, we need to compute the negative completed log-likelihood (here scaled by $n^{-1}$), namely the negative joint distribution of $\by$, $\bdelta$ and $\bz = (z_1, \ldots, z_n)^\top$. It can be written
\begin{align}
\ell_n^{\text{comp}}(\theta)&=\ell_n^{\text{comp}}(\theta ; \by, \bdelta, \bz) \nonumber \\ 
&= - n^{-1} \sum_{i=1}^n \Big \{ \delta_i \Big [ \sum_{k=0}^{K-1} \mathds{1}_{\{z_i = k\}} \big( \log \pi_{\beta_k}(x_i) + \log f_k(y_i;\alpha_k) \big) + \log \Gb(y_i^-) \Big ]  \nonumber \\
&\quad \quad \ \ + (1 - \delta_i) \Big [ \sum_{k=0}^{K-1} \mathds{1}_{\{z_i = k\}} \big( \log \pi_{\beta_k}(x_i) + \log \Fb_k(y_i^-;\alpha_k) \big) + \log g(y_i) \Big ] \Big \}. \label{eq:comp-lik}
\end{align}
Suppose that we are at step $l+1$ of the algorithm, with current iterate denoted $\theta^{(l)}=~(\alpha_0^{(l)}, \ldots, \alpha_{K-1}^{(l)}, \beta_0^{(l)}, \ldots, \beta_{K-1}^{(l)})^\top$. For the E-step, we need to compute the expected log-likelihood given by
\begin{equation*}
Q_n(\theta, \theta^{(l)}) = \E_{\theta^{(l)}}[\ell_n^{\text{comp}}(\theta) | \by, \bdelta].
\end{equation*}
We note that 
\begin{equation}
  \label{eq:qi}
  q_{i, k}^{(l)} = \E_{\theta^{(l)}} [\mathds{1}_{\{z_i = k\}} | y_i,\delta_i] 
  = \bP_{\theta^{(l)}} [z_i = k| y_i,\delta_i] 
  = \frac{\Lambda_{k,i}^{(l)}}{\sum_{r=0}^{K-1} \Lambda_{r,i}^{(l)}}
\end{equation}
with 
\begin{equation}
\Lambda_{k,i}^{(l)} = \big[ f_k(y_i;\alpha_k^{(l)})\Gb(y_i^-) \big]^{\delta_i}\big[ g(y_i)\Fb_k(y_i^-;\alpha_k^{(l)}) \big]^{1-\delta_i} \pi_{\beta_k^{(l)}}(x_i)  \label{eq:Lambda_k}
\end{equation}
so that $Q_n(\theta, \theta^{(l)})$ is obtained from~\eqref{eq:comp-lik} by replacing the two $\mathds{1}_{\{z_i = k\}}$ occurrences with $q_{i, k}^{(l)}$.
Depending on the chosen distributions $f_k$, the M-step can either be explicit for the updates of $\alpha_k$ (see Section~\ref{parameterization} below for the geometric distributions case), or obtained using a minimization algorithm otherwise.

Let us focus now on the update of $\beta_k$ in the M-step of the algorithm. By denoting
\begin{equation*}
  R_{n,k}^{(l)}(\beta_k) = -n^{-1} \sum_{i=1}^n q_{i, k}^{(l)} \log \pi_{\beta_k}(x_i)
\end{equation*}
the quantities involved in $Q_n$ that depend on $\beta_k$, the update for $\beta_k$ therefore requires to minimize
\begin{equation}
  \label{eq:sub-objective}
  R_{n, k}^{(l)}(\beta_k) + \gamma_k \big( (1-\eta)\norm{\beta_k}_1 + \frac \eta 2 \norm{\beta_k}_2^2 \big).
\end{equation}
The minimization Problem~\eqref{eq:sub-objective} is a convex problem. It looks like the logistic regression objective, where labels are not fixed but softly encoded by the expectation step (computation of $q_{i,k}^{(l)}$ above, see Equation~\eqref{eq:qi}).

We minimize~\eqref{eq:sub-objective} using the well-known L-BFGS-B algorithm \citep{zhu1997algorithm}. This algorithm  belongs to the class of quasi-Newton optimization routines, which solve the given minimization problem by computing approximations of the inverse Hessian matrix of the objective
function. It can deal with differentiable convex objectives with box constraints. In order to use it with $\ell_1$ penalization, which is not differentiable, we use the trick borrowed from \citet{andrew2007scalable}: for $a \in \R$, write $|a| = a_+ + a_-$, where $a_+$ and $a_-$ are respectively the positive and negative part of $a$, and add the constraints $a_+ \geq 0$ and $a_- \geq 0$.
Namely, we rewrite the minimization problem~\eqref{eq:sub-objective} as the following differentiable problem with box constraints
\begin{equation}
  \label{eq:sub-problem}
  \begin{split}
    \text{minimize}& \quad \quad R_{n, k}^{(l)}(\beta_k^+ - \beta_k^-) + \gamma_k (1-\eta) \sum_{j=1}^d (\beta_{k,j}^+ + \beta_{k,j}^-) + \gamma_k \frac \eta 2 \norm{\beta_k^+ - \beta_k^-}_2^2 \\
    \text{subject to}& \quad \quad \beta_{k,j}^+ \geq 0 \text{ and }\beta_{k,j}^- \geq 0 \text{ for all } j \in  \{1, \dots , d\},
  \end{split} 
\end{equation}
where $\beta_k^\pm = (\beta_{k,1}^\pm, \ldots, \beta_{k,d}^\pm)^\top$. The L-BFGS-B solver requires the exact value of the gradient, which is easily given by
\begin{equation}
  \label{eq:grad}
  \frac{\partial R_{n,k}^{(l)}(\beta_k)}{\partial \beta_k} = -n^{-1} \sum_{i=1}^n q_{i,k}^{(l)} \big( 1 - \pi_{\beta_k}(x_i) \big) x_i.
\end{equation}
In Algorithm~\ref{alg:QNEM}, we describe the main steps of the QNEM algorithm to minimize the function given in Equation~\eqref{eq:objective}.
 \begin{algorithm}
 \caption{\textbf{Algorithm 1: }QNEM Algorithm for inference of the C-mix model
     \label{alg:QNEM}}
 \begin{algorithmic}[1]
 \Require Training data $(x_i,y_i,\delta_i)_{i \in \{ 1, \dots, n \}}$; starting parameters ($\alpha_k^{(0)}, \beta_k^{(0)})_{k \in \{ 0, \dots, K-1 \}}$; tuning parameters $\gamma_k \geq 0$. 
 \vspace{.2cm}
 \For{$l=0, \ldots, $ until convergence }
 \State Compute $(q_{i, k}^{(l)})_{k \in \{ 0, \dots, K-1 \}}$ using Equation~\eqref{eq:qi}.
 \State Compute $(\alpha_k^{(l+1)})_{k \in \{ 0, \dots, K-1 \}}$.
 \State Compute $(\beta_k^{(l+1)})_{k \in \{ 0, \dots, K-1 \}}$ by solving~\eqref{eq:sub-problem} with the L-BFGS-B algorithm.
 \EndFor \\
 \Return Last parameters $(\alpha_k^{(l)},\beta_k^{(l)})_{k \in \{ 0, \dots, K-1 \}}$.
 \end{algorithmic}
 \end{algorithm}
 
\noindent The penalization parameters $\gamma_k$ are chosen using cross-validation, see Section~\ref{sec:Numerical details} of Appendices for precise statements about this procedure and about other numerical details.

\subsection{Convergence to a stationary point \label{Convergence}}
We are addressing here convergence properties of the QNEM algorithm described in Section~\ref{QNEM algorithm} for the minimization of the objective function defined in Equation~\eqref{eq:objective}. Let us denote 
\begin{equation*}
Q_n^{\text{pen}}(\theta, \theta^{(l)}) = Q_n(\theta, \theta^{(l)}) + \sum_{k=0}^{K-1} \gamma_k \big( (1-\eta)\norm{\beta_k}_1 + \frac \eta 2 \norm{\beta_k}_2^2 \big).
\end{equation*}
Convergence properties of the EM algorithm in a general setting are well known, see \citet{wu1983convergence}. In the QNEM algorithm, since we only improve $Q_n^{\text{pen}}(\theta, \theta^{(l)})$ instead of a minimization of $Q_n(\theta, \theta^{(l)})$, we are not in the EM algorithm setting but in a so called generalized EM (GEM) algorithm setting \citep{dempster1977maximum}. 
For such an algorithm, we do have the descent property, in the sense that the criterion function given in Equation~\eqref{eq:objective} is reduced at each iteration, namely
\begin{equation*}
  \ell^{\text{pen}}_n(\theta^{(l+1)}) \leq \ell^{\text{pen}}_n(\theta^{(l)}).
\end{equation*}
Let us make two hypothesis.
\begin{hyp}\label{hyp:bounded}
The duration densities $f_k$ are such that $\ell^{\text{pen}}_n$ is bounded for all $\theta$.
\end{hyp}
\begin{hyp}\label{hyp:strictlyconvex}
$Q_n^{\text{pen}}(\theta, \theta^{(l)})$ is continuous in $\theta$ and $\theta^{(l)}$, and for any fixed $\theta^{(l)}$, $Q_n^{\text{pen}}(\theta, \theta^{(l)})$ is a convex function in $\theta$ and is strictly convex in each coordinate of $\theta$.
\end{hyp}
\noindent Under Hypothesis~\ref{hyp:bounded}, $l \mapsto \ell^{\text{pen}}_n(\theta^{(l)})$ decreases monotically to some finite limit. 
By adding Hypothesis~\ref{hyp:strictlyconvex}, convergence of the QNEM algorithm to a stationary point can be shown. In particular, the stationary point is here a local minimum.

\begin{theorem}\label{th:cvg}
Under Hypothesis~\ref{hyp:bounded} and~\ref{hyp:strictlyconvex}, and considering the QNEM algorithm for the criterion function defined in Equation~\eqref{eq:objective}, every cluster point $\bar{\theta}$ of the sequence $\{ \theta^{(l)}; l=0,1,2,\dots \}$ generated by the QNEM algorithm is a stationary point of the criterion function defined in Equation~\eqref{eq:objective}.
\end{theorem}
\noindent A proof is given in Section~\ref{sec:th1 proof} of Appendices.

\subsection{Parameterization}
\label{parameterization}

Let us discuss here the parametrization choices we made in the experimental part. First, in many applications - including the one addressed in Section~\ref{sec:application} - we are interested in identifying one subgroup of the population with a high risk of adverse event compared to the others. Then, in the following, we consider $Z \in \{ 0, 1\}$ where $Z=1$ means high-risk of early death and $Z=0$ means low risk. Moreover, in such a setting where $K=2$, one can compare the learned groups by the C-mix and the ones learned by the CURE model in terms of survival curves (see Figure~\ref{fig:AUC-survival}).

To simplify notations and given the constraint formulated in Equation~\ref{eq:sum-to-1}, we set $\beta_0 = 0$ and we denote $\beta = \beta_1$ and $\pi_\beta(x)$ the conditional probability that a patient belongs to the group with high risk of death, given its covariates $x$.

In practice, we deal with discrete times in days. It turns out that the times of the data used for applications in Section~\ref{sec:application} is well fitted by Weibull distributions. This choice of distribution is very popular in survival analysis, see for instance~\citet{klein2005survival}. We then first derive the QNEM algorithm with 
\begin{equation*}
f_k(t;\alpha_k) = (1 - \phi_k)^{t^{\mu_k}} - (1 - \phi_k)^{{(t+1)}^{\mu_k}}
\end{equation*}
with here $\alpha_k = (\phi_k,\mu_k) \in (0, 1) \times \R_+$, $\phi_k$ being the scale parameter and $\mu_k$ the shape parameter of the distribution.

As explained in the following Section~\ref{sec:Performance evaluation}, we select the best model using a cross-validation procedure based on the C-index metric, and the performances are evaluated according to both C-index and AUC($t$) metrics (see Sections~\ref{sec:Metrics} for details). Those two metrics have the following property: if we apply any mapping on the marker vector (predicted on a test set) such that the order between all vector coefficient values is conserved, then both C-index and AUC($t$) estimates remain unchanged. In other words, by denoting $(M_i)_{i \in \{1,\dots,n_{\text{test}}\}}$ the vector of markers predicted on a test set of $n_{\text{test}}$ individuals, if $\psi$ is a function such that for all $ (i,j) \in \{1,\dots,n_{\text{test}}\}^2, \big( M_i < M_j \Rightarrow \psi(M_i) < \psi(M_j) \big)$, then both C-index and AUC($t$) estimates induced by $(M_i)_{i \in \{1,\dots,n_{\text{test}}\}}$ or by $\big( \psi(M_i) \big)_{i \in \{1,\dots,n_{\text{test}}\}}$ are the same.

The order in the marker coefficients is actually paramount when the performances are evaluated according to the mentioned metrics. Furthermore, it turns out that empirically, if we add a constraint on the mixture of Weibull that enforces an \textit{order like} relation between the two distributions $f_0$ and $f_1$, the performances are improved. To be more precise, the constraint to impose is that the two density curves do not intersect. We then choose to impose the following: the two scale parameters are equal, $i.e.\ \phi_0 = \phi_1 = \phi$. Indeed under this hypothesis, we do have that for all $\phi \in (0, 1),\  \big( \mu_0 < \mu_1 \Rightarrow \forall t \in \R_+, f_0(t;\alpha_0) > f_1(t;\alpha_1) \big)$.

With this Weibull parameterization, updates for $\alpha_k$ are not explicit in the QNEM algorithm, and consequently require some iterations of a minimization algorithm. Seeking to have explicit updates for $\alpha_k$, we then derive the algorithm with geometric distributions instead of Weibull (geometric being a particular case of Weibull with $\mu_k=1$), namely $f_k(t;\alpha_k) = \alpha_k (1-\alpha_k)^{t-1}$ with $\alpha_k \in (0, 1)$. 

With this parameterization, we obtain from Equation~\eqref{eq:Lambda_k}
\begin{align*}
\Lambda_{1,i}^{(l)} &= \big[ \alpha_1^{(l)}(1-\alpha_1^{(l)})^{y_i-1} \big]^{\delta_i} \big[ (1-\alpha_1^{(l)})^{y_i} \big]^{1-\delta_i} \pi_{\beta^{(l)}}(x_i) \quad \text{and} \\
\Lambda_{0,i}^{(l)} &= \big[ \alpha_0^{(l)}(1-\alpha_0^{(l)})^{y_i-1} \big]^{\delta_i} \big[ (1-\alpha_0^{(l)})^{y_i} \big]^{1-\delta_i} \big(1-\pi_{\beta^{(l)}}(x_i)\big),
\end{align*}
which leads to the following explicit M-step
\begin{equation*}
  \alpha_0^{(l+1)} = \frac{\sum_{i=1}^n \delta_i(1 - q_{i}^{(l)})}{\sum_{i=1}^n (1 - q_{i}^{(l)}) y_i } \;\; \text{ and } \;\;
  \alpha_1^{(l+1)} = \frac{\sum_{i=1}^n \delta_i q_{i}^{(l)}}{\sum_{i=1}^n q_{i}^{(l)} y_i}.
\end{equation*}
In this setting, implementation is hence straightforward. Note that Hypothesis~\ref{hyp:bounded} and~\ref{hyp:strictlyconvex} are immediately satisfied with this geometric parameterization.

In Section~\ref{sec:application}, we note that performances are similar for the C-mix model with Weibull or geometric distributions on all considered biomedical datasets. The geometric parameterization leading to more straightforward computations, it is the one used to parameterize the C-mix model in what follows, if not otherwise stated. Let us focus now on the performance evaluation of the C-mix model and its comparison with the Cox PH and CURE models, both regularized with the Elastic-Net.

\section{Performance evaluation\label{sec:Performance evaluation}}
In this section, we first briefly introduce the models we consider for performance comparisons. Then, we provide details regarding the simulation study and data generation. The chosen metrics for evaluating performances are then presented, followed by the results.

\subsection{Competing models\label{sec:competing models}}

The first model we consider is the Cox PH model penalized by the Elastic-Net, denoted Cox PH in the following.
In this model introduced in \citet{Cox1972JRSS}, the partial log-likelihood is given by
\begin{equation*}
  \ell_n^{\text{cox}}(\beta) = n^{-1} \sum_{i=1}^n \delta_i \big( x_i^\top \beta - \log \sum_{i' : y_{i'} \geq y_i} \text{exp}(x_{i'}^\top \beta) \big).
\end{equation*}
We use respectively the \texttt{R} packages \texttt{survival} and \texttt{glmnet} \citep{simon2011regularization} for the partial log-likelihood and the minimization of the following quantity
\begin{equation*}
  - \ell_n^{\text{cox}}(\beta) + \gamma \big( (1-\eta)\norm{\beta}_1 + \frac \eta 2 \norm{\beta}_2^2 \big),
\end{equation*}
where $\gamma$ is chosen by the same cross-validation procedure than the C-mix model, for a given $\eta$ (see Section~\ref{sec:Numerical details} of Appendices. Ties are handled via the Breslow approximation of the partial likelihood \citep{breslow1972contribution}.

We remark that the model introduced in this paper cannot be reduced to a Cox model. 
Indeed, the C-mix model intensity can be written (in the geometric case)
\begin{equation*}
\lambda(t)= \dfrac{\alpha_1(1-\alpha_1)^{t-1} + \alpha_0(1-\alpha_0)^{t-1}\ \exp(x^\top \beta) }{(1-\alpha_1)^t + (1-\alpha_0)^t\ \text{exp}(x^\top \beta)},
\end{equation*}
while it is given by Equation~\eqref{eq:cox-intensity} in the Cox model.

Finally, we consider the CURE~\cite{farewell1982use} model penalized by the Elastic-Net and denoted CURE in the following, with a logistic function for the incidence part and a parametric survival model for $S(t|Z=1)$, where $Z=0$ means that patient is cured, $Z=1$ means that patient is not cured, and $S(t) = \exp(-\int_0^t\lambda(s) \dd s)$ denotes the survival function. In this model, we then have $S(t|Z=0)$ constant and equal to 1. We add an Elastic-Net regularization term, and since we were not able to find any open source package where CURE models were implemented with a regularized objective, we used the QNEM algorithm in the particular case of CURE model.
We just add the constraint that the geometric distribution $\cG(\alpha_0)$ corresponding to the cured group of patients ($Z=0$) has a parameter $\alpha_0=0$, which does not change over the algorithm iterations. 
The QNEM algorithm can be used in this particular case, were some terms have disapeared from the completed log-likelihood, since in the CURE model case we have $\big\{i \in \{1, \dots , n\} : z_i=0,\ \delta_i = 1 \big\} = \varnothing$. Note that in the original introduction of the CURE model in~\citet{farewell1982use}, the density of uncured patients directly depends on individual patient covariates, which is not the case here.

We also give additional simulation settings in Section~\ref{sec:additional comparison} of Appendices. First, the case where $d \gg n$, including a comparison of the screening strategy we use in Section~\ref{sec:application} with the iterative sure independence screening~\citep{fan2010high} (ISIS) method. We also add  simulations where data is generated according to the C-mix model with gamma distributions instead of geometric ones, and include the accelerated failure time model~\citep{wei1992accelerated} (AFT) in the performances comparison study.

\subsection{Simulation design}
\label{simulation design}
In order to assess the proposed method, we perform an extensive Monte Carlo simulation study.
Since we want to compare the performances of the 3 models mentioned above, we consider 3 simulation cases for the time distribution: one for each competing model. We first choose a coefficient vector $\beta=~(\underbrace{\nu,\ldots,\nu}_s,0,\ldots,0) \in \R^d$, with $\nu\in\R$ being the value of the active coefficients and $s\in \{1,\dots,d\}$ a sparsity parameter. For a desired low-risk patients proportion $\pi_0 \in [0,1]$, the high-risk patients index set is given by
\begin{equation*}
  \cH = \big\{ \lfloor (1-\pi_0) \times n \rfloor \;\; \text{random samples without replacement} \big\} \subset \{1, \dots, n \},
\end{equation*}
where $\lfloor a \rfloor$ denotes the largest integer less than or equal to $a \in \R$. 
For the generation of the covariates matrix, we first take $[x_{ij}] \in \R^{n \times d} \sim \cN(0,\bSigma(\rho))$, with $\bSigma(\rho)$ a $(d \times d)$ Toeplitz covariance matrix \citep{mukherjee1988some} with correlation $\rho \in (0, 1)$. 
We then add a gap $\in \R^+$ value for patients $i \in \cH$ and subtract it for patients $i \notin \cH$, only on active covariates plus a proportion $r_{cf}\in [0,1]$ of the non-active covariates considered as confusion factors, that is
\begin{center}
$x_{ij} \leftarrow x_{ij} \pm \text{gap}\ \text{for } j \in \big\{1, \dots, s, \dots, \lfloor (d-s)r_{cf} \rfloor \big\}$.
\end{center}
Note that this is equivalent to generate the covariates according to a gaussian mixture.

Then we generate $Z_i \sim \cB\big(\pi_\beta(x_i)\big)$ in the C-mix or CURE simulation case, where $\pi_\beta(x_i)$ is computed given Equation~\eqref{eq:pi}, with geometric distributions for the durations (see Section~\ref{parameterization}). 
We obtain $T_i \sim \cG(\alpha_{Z_i})$ in the C-mix case, and 
$T_i \sim \infty \mathds{1}_{\{Z_i=0\}} + \cG(\alpha_1) \mathds{1}_{\{Z_i=1\}}$ in the CURE case. For the Cox PH model, we take $T_i \sim - \log(U_i)\exp(-x_i^\top \beta)$, with $U_i \sim \cU([0,1])$ and where $\cU([a,b])$ stands for the uniform distribution on a segment $[a,b]$. 

The distribution of the censoring variable $C_i$ is geometric $\cG(\alpha_c)$, with $\alpha_c \in (0, 1)$. 
The parameter $\alpha_c$ is tuned to maintain a desired censoring rate $r_c \in [0,1]$, using a formula given in Section~\ref{sec:censoring level} of Appendices.
The values of the chosen hyper parameters are sumarized in Table~\ref{table:parameters choice}. 
\begin{table}[!htb]
\centering
\caption{Hyper-parameters choice for simulation}
\resizebox{\textwidth}{!}{
\begin{tabular}{|c|c|c|c|c|c|c|c|c|c|c|c|}
\hline
$\eta$ & $n$ & $d$ & $s$ & $r_{cf}$ & $\nu$ & $\rho$ & $\pi_0$ & gap & $r_c$ & $\alpha_0$ & $\alpha_1$  \\
\hline
0.1 & 100, 200, 500 & 30, 100 & 10 & 0.3 & 1 & 0.5 & 0.75 & 0.1, 0.3, 1 & 0.2, 0.5 & 0.01 & 0.5 \\
\hline
\end{tabular}}
\label{table:parameters choice}
\end{table}

\noindent Note that when simulating under the CURE model, the proportion of censored time events is at least equal to $\pi_0$ : we then choose $\pi_0 = 0.2$ for the CURE simulations only.

Finally, we want to assess the stability of the C-mix model in terms of variable selection and compare it to the CURE and Cox PH models. To this end, we follow the same simulation procedure explained in the previous lines. For each simulation case, we make vary the two hyper-parameters that impact the most the stability of the variable selection, that is the gap varying in $[0,2]$ and the confusion rate $r_{cf}$ varying in $[0,1]$. All other hyper-parameters are the same than in Table~\ref{table:parameters choice}, except $s=150$ and with the choice $(n,d)=(200,300)$. For a given hyper-parameters configuration (gap, $r_{cf}$), we use the following approach to evaluate the variable selection power of the models. Denoting $\tilde{\beta}_i = |\hat{\beta}_i| / \text{max} \big\{ |\hat{\beta}_i|, i \in \{1,\ldots,d \} \big\}$, if we consider that $\tilde{\beta}_i$ is the predicted probability that the true $\beta_i$ equals $\nu$, then we are in a binary prediction setting and we use the resulting AUC of this problem. Explicit examples of such AUC computations are given in Section~\ref{sec:details feature selection} of Appendices.

\subsection{Metrics\label{sec:Metrics}}
We detail in this section the metrics considered to evaluate risk prediction performances. Let us denote by $M$ the marker under study. Note that $M=\pi_{\hat{\beta}}(X)$ in the C-mix and the CURE model cases, and $M=\exp(X^\top \hat{\beta}^{\text{cox}})$ in the Cox PH model case. We denote by $h$ the probability density function of marker $M$, and assume that the marker is measured once at $t = 0$.

For any threshold $\xi$, cumulative true positive rates and dynamic false positive rates are two functions of time respectively defined as $\text{TPR}^\C(\xi,t)=\bP[M>\xi | T \leq t]$ and $\text{FPR}^\D(\xi,t)=\bP[M>\xi | T > t]$. Then, as introduced in \citet{heagerty2000time}, the cumulative dynamic time-dependent AUC is defined as follows
\begin{align*}
  \text{AUC}^{\C,\D}(t) &= \int_{-\infty}^{\infty} \text{TPR}^\C(\xi,t) \left|\frac{\partial \text{FPR}^\D(\xi,t)}{\partial \xi}\right| \dd \xi,
\end{align*}
that we simply denote AUC($t$) in the following.
We use the Inverse Probability of Censoring Weighting (IPCW) estimate of this quantity with a Kaplan-Meier estimator of the conditional survival function $\bP[T>t | M=m ]$, as proposed in \citet{blanche2013time} and already implemented in the \texttt{R} package \texttt{timeROC}.

A common concordance measure that does not depend on time is the C-index~\citep{harrell1996tutorial} defined by
\begin{equation*}
  \cC =\bP[M_i > M_j | T_i < T_j],
\end{equation*}
with $i \neq j$ two independent patients (which does not depend on $i, j$ under the i.i.d. sample hypothesis). 
In our case,  $T$ is subject to right censoring, so one would typically consider the modified $\cC_\tau$ defined by
\begin{equation*}
  \cC_\tau =\bP[M_i > M_j | Y_i < Y_j , Y_i < \tau],
\end{equation*}
with $\tau$ corresponding to the fixed and prespecified follow-up period duration \citep{heagerty2005survival}. A Kaplan-Meier estimator for the censoring distribution leads to a nonparametric and consistent estimator of $\cC_\tau$ \citep{uno2011c}, already implemented in the \texttt{R} package \texttt{survival}.

Hence in the following, we consider both AUC($t$) and C-index metrics to assess performances.

\subsection{Results of simulation}
\label{Results}
\begin{figure}[!h]
\begin{center}
\centerline{\includegraphics[width=15cm]{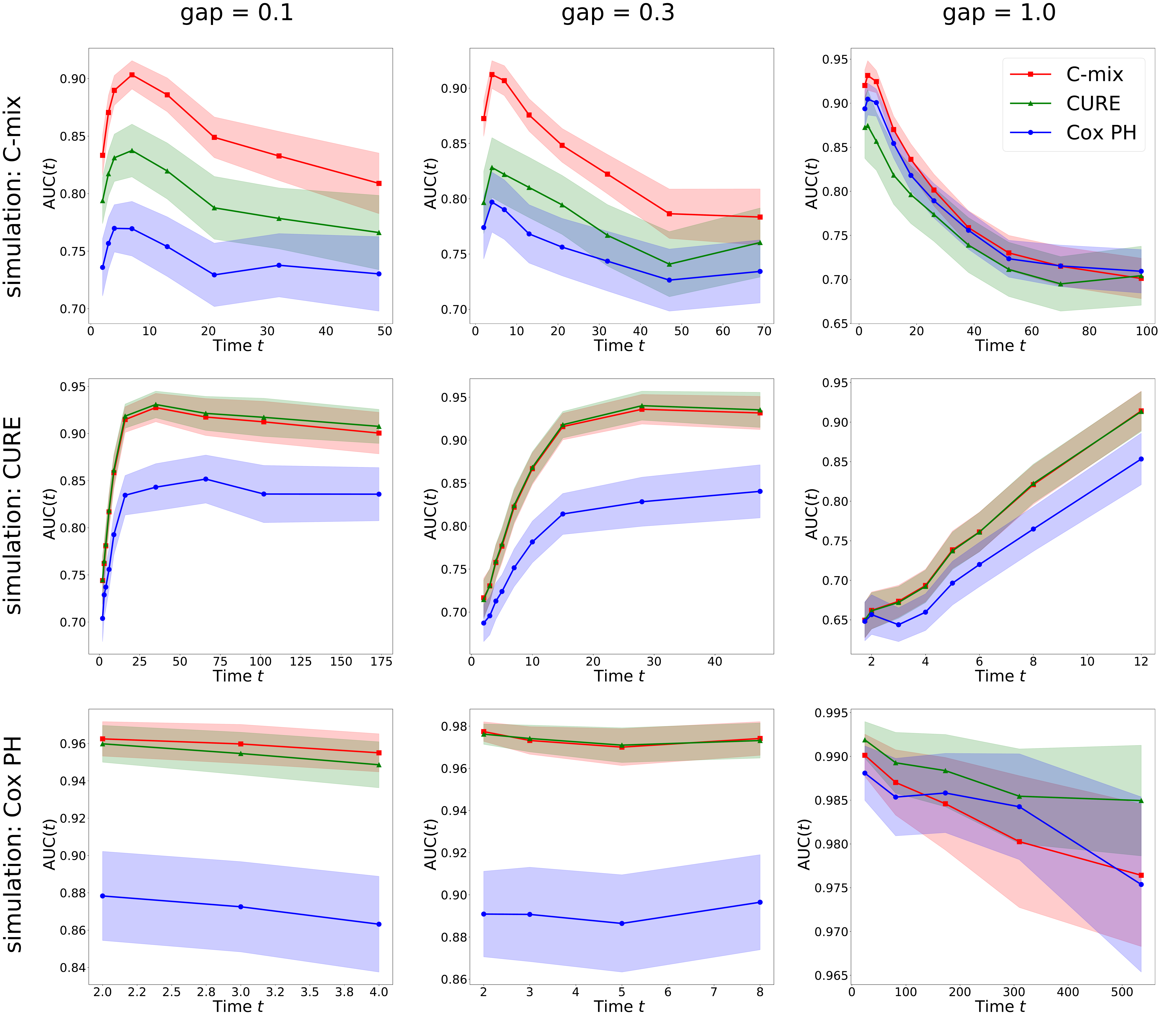}}
\end{center}
\captionsetup{justification=justified}
\caption{Average (bold lines) and standard deviation (bands) for $\text{AUC}(t)$ on 100 simulated data with $n=100$, $d=30$ and $r_c=0.5$. Rows correspond to the model simulated (cf. Section~\ref{simulation design}) while columns correspond to different gap values (the problem becomes more difficult as the gap value decreases). Surprisingly, our method gives almost always the best results, even under model misspecification (see Cox PH and CURE simulation cases on the second and third rows).
\label{figure: AUC(t)}}
\end{figure}

We present now the simulation results concerning the C-index metric in the case $(d, r_c) = (30, 0.5)$ in Table~\ref{table:C-index d=30}.  
See Section~\ref{sec:simulation results} of Appendices for results on other configurations for $(d,r_c)$.
Each value is obtained by computing the C-index average and standard deviation (in parenthesis) over 100 simulations.
The $\text{AUC}(t)$ average (bold line) and standard deviation (bands) over the same 100 simulations are then given in Figure~\ref{figure: AUC(t)}, where $n=100$.
Note that the value of the gap can be viewed as a difficulty level of the problem, since the higher the value of the gap, the clearer the separation between the two populations (low risk and high risk patients).

The results measured both by AUC($t$) and C-index lead to the same conclusion: the C-mix model almost always leads to the best results, even under model misspecification, $i.e.$ when data is generated according to the CURE or Cox PH model.
Namely, under CURE simulations, C-mix and CURE give very close results, with a strong improvement over Cox PH. 
Under Cox PH and C-mix simulations, C-mix outperforms both Cox PH and CURE. 
Surprisingly enough, this exhibits a strong generalization property of the C-mix model, over both Cox PH and CURE.
Note that this phenomenon is particularly strong for small gap values, while with an increasing gap (or an increasing sample size $n$), all procedures barely exhibit the same performance.
It can be first explained by the non parametric baseline function in the Cox PH model, and second by the fact that unlike the Cox PH model, the C-mix and CURE models exploit directly the mixture aspect.

\begin{figure}[!h]
\begin{center}
\centerline{\includegraphics[width=15cm]{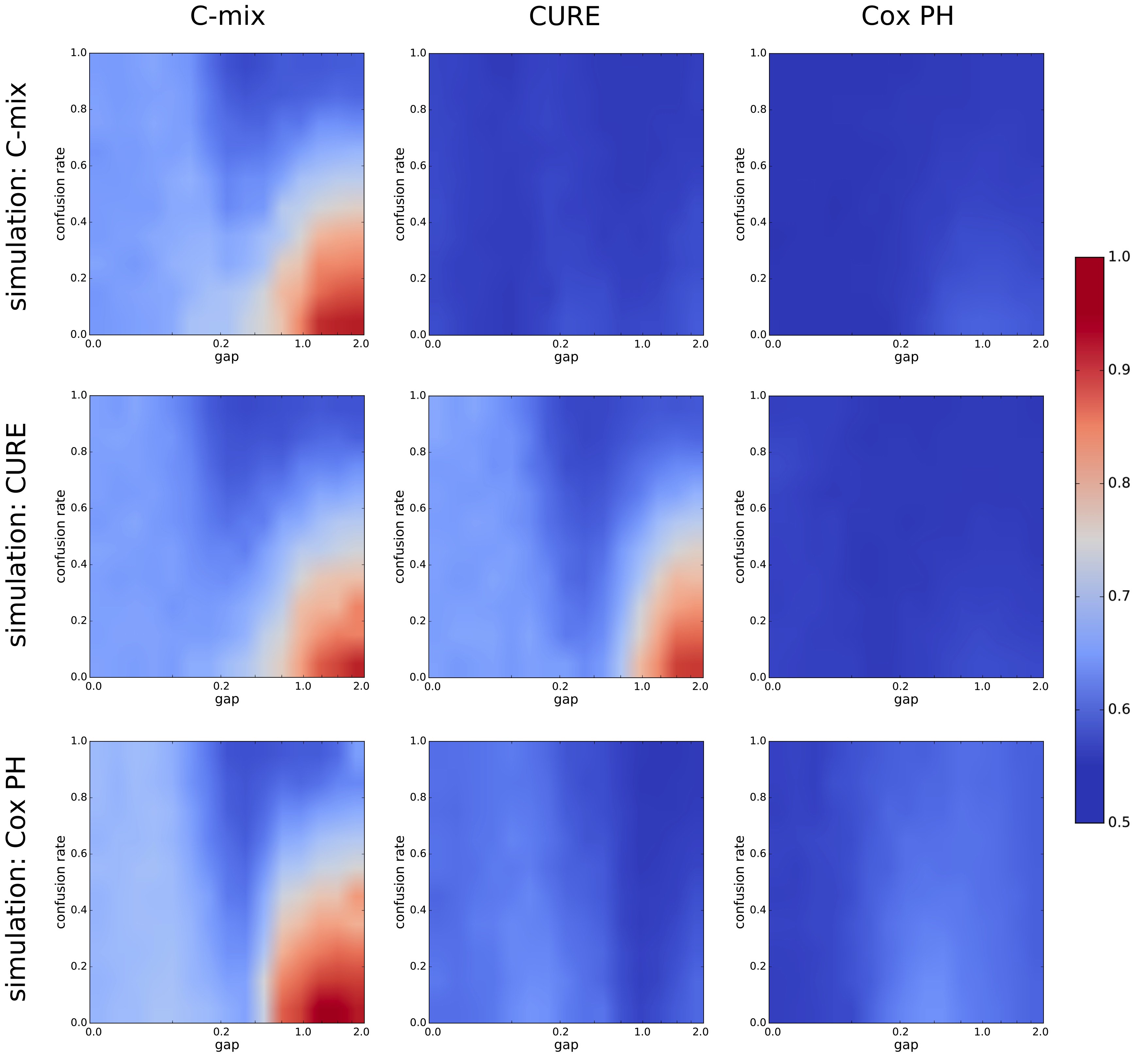}}
\end{center}
\captionsetup{justification=justified}
\caption{Average AUC calculated according to Section~\ref{simulation design} and obtained after 100 simulated data for each (gap, $r_{cf}$) configuration (a grid of 20x20 different configurations is considered). A gaussian interpolation is then performed to obtain smooth figures. Note that the gap values are $log$-scaled. Rows correspond to the model simulated while columns correspond to the model under consideration for the variable selection evaluation procedure. Our method gives the best results in terms of variable selection, even under model misspecification.
\label{figure: feature selection}}
\end{figure}

Finally, Figure~\ref{figure: feature selection} gives the results concerning the stability of the variable selection aspect of the competing models.
The C-mix model appears to be the best method as well considering the variable selection aspect, even under model misspecification. We notice a general behaviour of our method that we describe in the following, which is also shared by the CURE model only when the data is simulated according to itself, and which justifies the log scale for the gap to clearly distinguish the three following phases.
For very small gap values (less than $0.2$), the confusion rate $r_{cf}$ does not impact the variable selection performances, since adding very small gap values to the covariates is almost imperceptible. This means that the resulting AUC is the same when there is no confusion factors and when $r_{cf}=1$ (that is when there are half active covariates and half confusion ones).
For medium gap values (saying between $0.2$ and 1), the confusion factors are more difficult to identify by the model as there number goes up (that is when $r_{cf}$ increases), which is precisely the confusion factors effect we expect to observe. 
Then, for large gap values (more than 1), the model succeeds in vanishing properly all confusion factors since the two subpopulations are more clearly separated regarding the covariates, and the problem becomes naturally easier as the gap increases.
\begin{sidewaystable}
\centering
\captionsetup{justification=justified}
\caption{Average C-index on 100 simulated data and standard deviation in parenthesis, with $d=30$ and $r_c=0.5$. For each configuration, the best result appears in bold.}
\resizebox{\textwidth}{!}{%
\begin{tabular}{ccccccccccc}
\toprule
& & \multicolumn{9}{c@{}}{$\textnormal{Estimation}$}\\
\cmidrule(l){3-11}
& & \multicolumn{3}{c@{}}{ \hspace{-.7cm} $n=100$ \hspace{.7cm}} & \multicolumn{3}{c@{}}{\hspace{-1.2cm} $n=200$ \hspace{.7cm}} & \multicolumn{3}{c@{}}{$n=500$} \\
Simulation & gap \hspace{.8cm} & C-mix & CURE & \hspace{-.5cm} Cox\ PH \hspace{.5cm} & C-mix & CURE & \hspace{-.5cm} Cox\ PH \hspace{.5cm} & C-mix & CURE & Cox\ PH \\
\midrule
 & 0.1 \hspace{.8cm} & \textbf{0.786 (0.057)} & 0.745 (0.076) & 0.701 (0.075) \hspace{.8cm} & \textbf{0.792 (0.040)} & 0.770 (0.048) & 0.739 (0.055) \hspace{.8cm} & \textbf{0.806 (0.021)} & 0.798 (0.023) & 0.790 (0.024) \\
C-mix & 0.3 \hspace{.8cm} & \textbf{0.796 (0.055)} & 0.739 (0.094) & 0.714 (0.088) \hspace{.8cm} & \textbf{0.794 (0.036)} & 0.760 (0.058) & 0.744 (0.055) \hspace{.8cm} & \textbf{0.801 (0.021)} & 0.784 (0.027) & 0.783 (0.026) \\
 & 1 \hspace{.8cm} & \textbf{0.768 (0.062)} & 0.734 (0.084) & 0.756 (0.066) \hspace{.8cm} & \textbf{0.766 (0.043)} & 0.736 (0.054) & 0.764 (0.042) \hspace{.8cm} & 0.772 (0.026) & 0.761 (0.027) & \textbf{0.772 (0.025)} \\
 \midrule
 & 0.1 \hspace{.8cm} & 0.770 (0.064) & \textbf{0.772 (0.062)} & 0.722 (0.073) \hspace{.8cm} & \textbf{0.790 (0.038)} & \textbf{0.790 (0.038)} & 0.758 (0.049) \hspace{.8cm} & 0.798 (0.025) & \textbf{0.799 (0.024)} & 0.787 (0.025) \\
CURE & 0.3 \hspace{.8cm} & \textbf{0.733 (0.073)} & 0.732 (0.072) & 0.686 (0.072) \hspace{.8cm} & 0.740 (0.053) & \textbf{0.741 (0.053)} & 0.714 (0.060) \hspace{.8cm} & \textbf{0.751 (0.029)} & \textbf{0.751 (0.029)} & 0.738 (0.030) \\
 & 1 \hspace{.8cm} & \textbf{0.659 (0.078)} & 0.658 (0.078) & 0.635 (0.070) \hspace{.8cm} & \textbf{0.658 (0.053)} & \textbf{0.658 (0.053)} & 0.647 (0.047) \hspace{.8cm} & \textbf{0.657 (0.031)} & \textbf{0.657 (0.031)} & 0.656 (0.032) \\
 \midrule
  & 0.1 \hspace{.8cm} & \textbf{0.940 (0.041)} & 0.937 (0.044) & 0.850 (0.097) \hspace{.8cm} & \textbf{0.959 (0.021)} & 0.958 (0.020) & 0.915 (0.042) \hspace{.8cm} & \textbf{0.964 (0.012)} & \textbf{0.964 (0.012)} & 0.950 (0.016) \\
Cox PH & 0.3 \hspace{.8cm} & \textbf{0.956 (0.030)} & 0.955 (0.029) & 0.864 (0.090) \hspace{.8cm} & \textbf{0.966 (0.020)} & 0.965 (0.020) & 0.926 (0.043) \hspace{.8cm} & 0.968 (0.013) & \textbf{0.969 (0.012)} & 0.959 (0.016) \\
 & 1 \hspace{.8cm} & 0.983 (0.016) & \textbf{0.985 (0.015)} & 0.981 (0.019) \hspace{.8cm} & 0.984 (0.012) & 0.985 (0.011) & \textbf{0.988 (0.010)} \hspace{.8cm} & 0.984 (0.007) & 0.985 (0.006) & \textbf{0.990 (0.005)} \\
\bottomrule
\end{tabular}}
\label{table:C-index d=30}
\end{sidewaystable}

\section{Application to genetic data \label{sec:application}}
In this section, we apply our method on three genetic datasets and compare its performance to the Cox PH and CURE models. We extracted normalized expression data and survival times $Y$ in days from breast invasive carcinoma (BRCA, $n=1211$), glioblastoma multiforme (GBM, $n=168$) and kidney renal clear cell carcinoma (KIRC, $n=605$). 

These datasets are available on The Cancer Genome Atlas (TCGA) platform, which aims at accelerating the understanding of the molecular basis of cancer through the application of genomic technologies, including large-scale genome sequencing. 
For each patient, 20531 covariates corresponding to the normalized gene expressions are available. 
We randomly split all datasets into a training set and a test set (30\% for testing, cross-validation is done on the training).

We compare the three models both in terms of C-index and AUC($t$) on the test sets.
Inference of the Cox PH model fails in very high dimension on the considered data with the \texttt{glmnet} package. 
We therefore make a first variable selection (screening) among the 20531 covariates.
To do so, we compute the C-index obtained by univariate Cox PH models (not to confer advantage to our method), namely Cox PH models fitted on each covariate separately. 
We then ordered the obtained 20531 C-indexes by decreasing order and extracted the top $d = 100$, $d = 300$ and $d = 1000$ covariates. We then apply the three methods on the obtained covariates.

\begin{figure}[!htb]
\centering
\captionsetup[subfigure]{justification=centering}
\begin{subfigure}{.33\textwidth}
  \centering
  \includegraphics[width=1\linewidth]{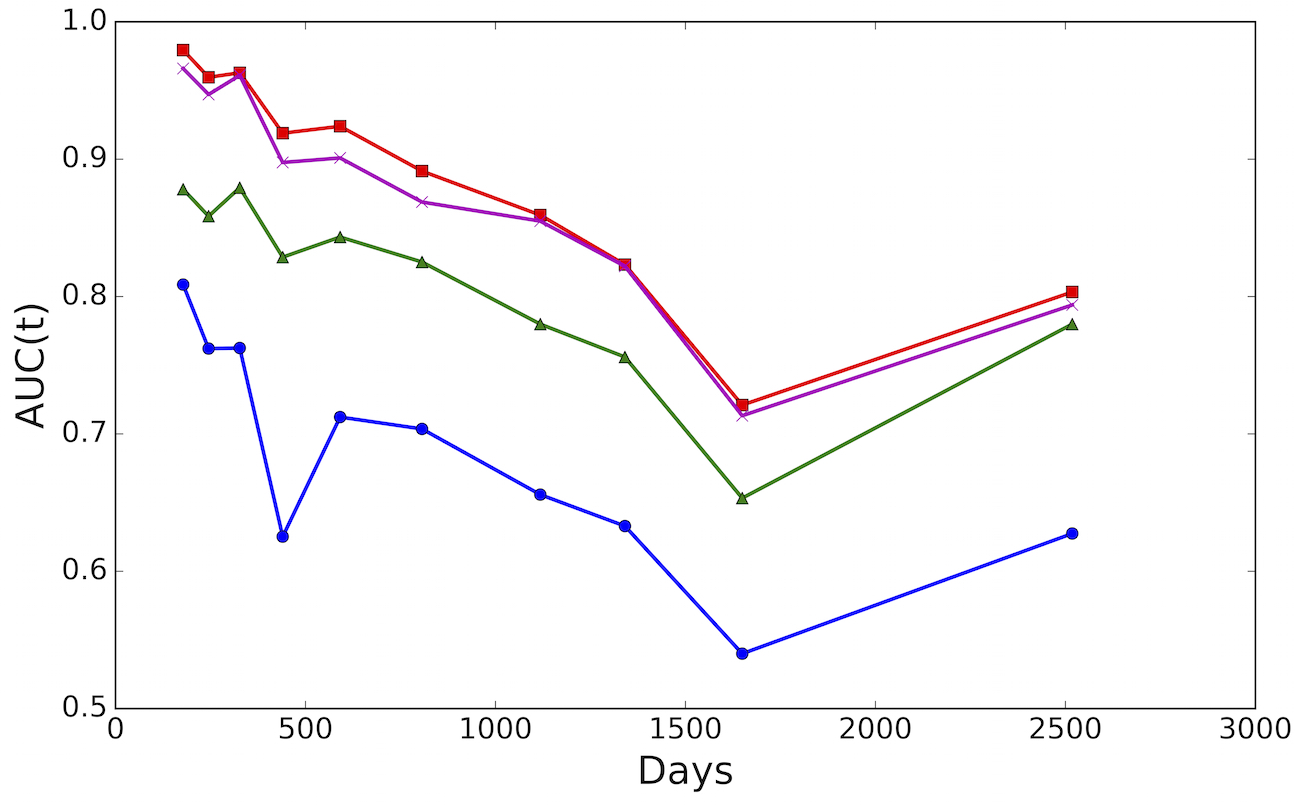}
  \caption{BRCA}
  \label{fig:sub1}
\end{subfigure}%
\begin{subfigure}{.33\textwidth}
  \centering
  \includegraphics[width=1\linewidth]{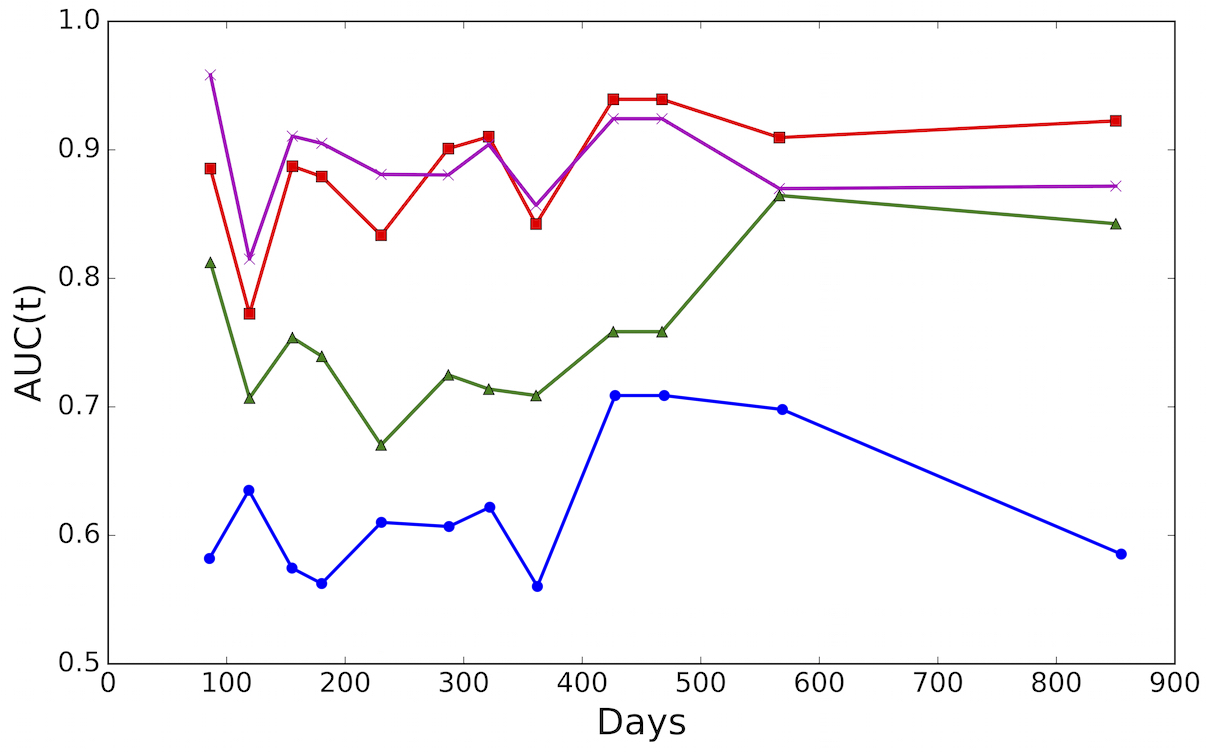}
  \caption{GBM}
  \label{fig:sub2}
\end{subfigure}
\begin{subfigure}{.33\textwidth}
  \centering
  \includegraphics[width=1\linewidth]{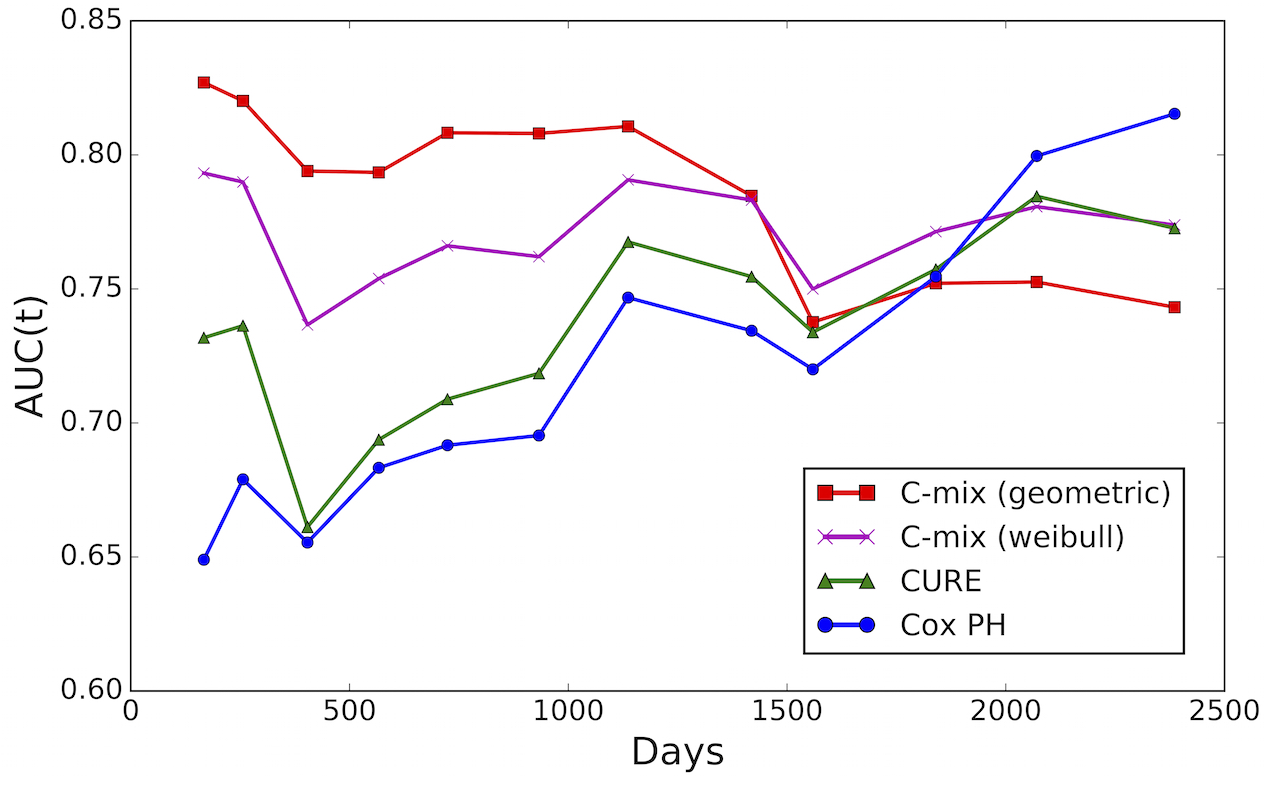}
  \caption{KIRC}
  \label{fig:sub2}
\end{subfigure}
\captionsetup{justification=justified}
\caption{$\text{AUC}(t)$ comparison on the three TCGA data sets considered, for $d=300$. We observe that C-mix model leads to the best results (higher is better) and outperforms both Cox PH and CURE in all cases. Results are similar in terms of performances for the C-mix model with geometric or Weibull distributions.}
\label{fig:AUC(t)}
\end{figure}

The results in terms of AUC($t$) curves are given in Figure~\ref{fig:AUC(t)} for $d = 300$, where we distinguish the C-mix model with geometric or Weibull distributions.

\noindent Then it appears that the performances are very close in terms of AUC($t$) between the C-mix model with geometric or Weibull distributions, which is also validated if we compare the corresponding C-index for these two parameterizations in Table~\ref{table:C-index C-mix comparison}. 

\begin{table}[!h]
\centering
\captionsetup{justification=justified}
\caption{C-index comparison between geometric or Weibull parameterizations for the C-mix model on the three TCGA data sets considered (with $d=300$). In all cases, results are very similar for the two distribution choices.}
\begin{tabular}{cccc}
\toprule
Parameterization & & Geometric & Weibull \\
\midrule
 & BRCA & 0.782 & 0.780\\
Cancer & GBM & 0.755 & 0.754 \\
 & KIRC & 0.849 & 0.835 \\
\bottomrule
\end{tabular}
\label{table:C-index C-mix comparison}
\end{table}
Similar conclusions in terms of C-index, AUC($t$) and computing time can be made on all considered datasets and for any choice of $d$. Hence, as already mentionned in Section~\ref{parameterization}, we only concentrate on the geometric parameterization for the C-mix model. 
The results in terms of C-index are then given in Table~\ref{table:C-index}.
\begin{table}[!htb]
\centering
\captionsetup{justification=justified}
\caption{C-index comparison on the three TCGA data sets considered. In all cases, C-mix gives the best results (in bold).}
\resizebox{\textwidth}{!}{%
\begin{tabular}{cccccccccccccc}
\toprule
Cancer & & \hspace{.2cm} & \multicolumn{3}{c@{}}{$\textnormal{BRCA}$} & \hspace{.2cm} & \multicolumn{3}{c@{}}{$\textnormal{GBM}$} & \hspace{.2cm} & \multicolumn{3}{c@{}}{$\textnormal{KIRC}$} \\
\cmidrule(l){4-6} \cmidrule(l){8-10} \cmidrule(l){12-14}
Model & & \hspace{.2cm} & C-mix & CURE & Cox PH & \hspace{.2cm} & C-mix & CURE & Cox PH & \hspace{.2cm} & C-mix & CURE & Cox PH \\
\midrule
 & 100 & \hspace{.2cm} & \textbf{0.792} & 0.764 & 0.705 & \hspace{.2cm} & \textbf{0.826} & 0.695 & 0.571 & \hspace{.2cm} & \textbf{0.768} & 0.732 & 0.716 \\
$d$ & 300 & \hspace{.2cm} & \textbf{0.782} & 0.753 & 0.723 & \hspace{.2cm} & \textbf{0.849} & 0.697 & 0.571 & \hspace{.2cm} & \textbf{0.755} & 0.691 & 0.698 \\
 & 1000 & \hspace{.2cm} & \textbf{0.817} & 0.613 & 0.577 & \hspace{.2cm} & \textbf{0.775} & 0.699 & 0.592 & \hspace{.2cm} & \textbf{0.743} & 0.690 & 0.685 \\
\bottomrule
\end{tabular}}
\label{table:C-index}
\end{table}

\begin{figure}[!htb]
\centering
\captionsetup[subfigure]{justification=centering}
\begin{subfigure}{.33\textwidth}
  \centering
  \includegraphics[width=1\linewidth]{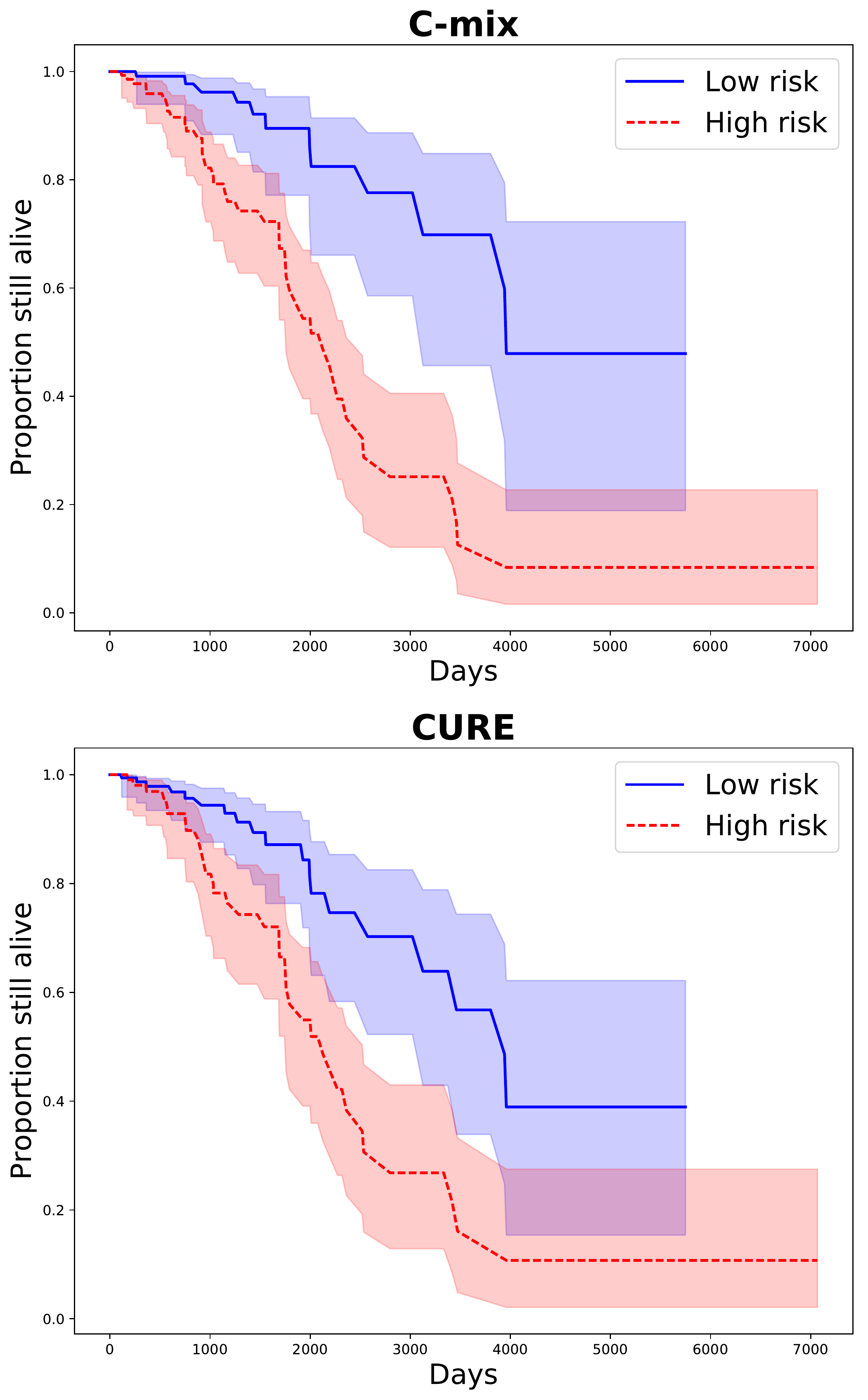}
  \caption{BRCA}
  \label{fig:sub1}
\end{subfigure}%
\begin{subfigure}{.33\textwidth}
  \centering
  \includegraphics[width=1\linewidth]{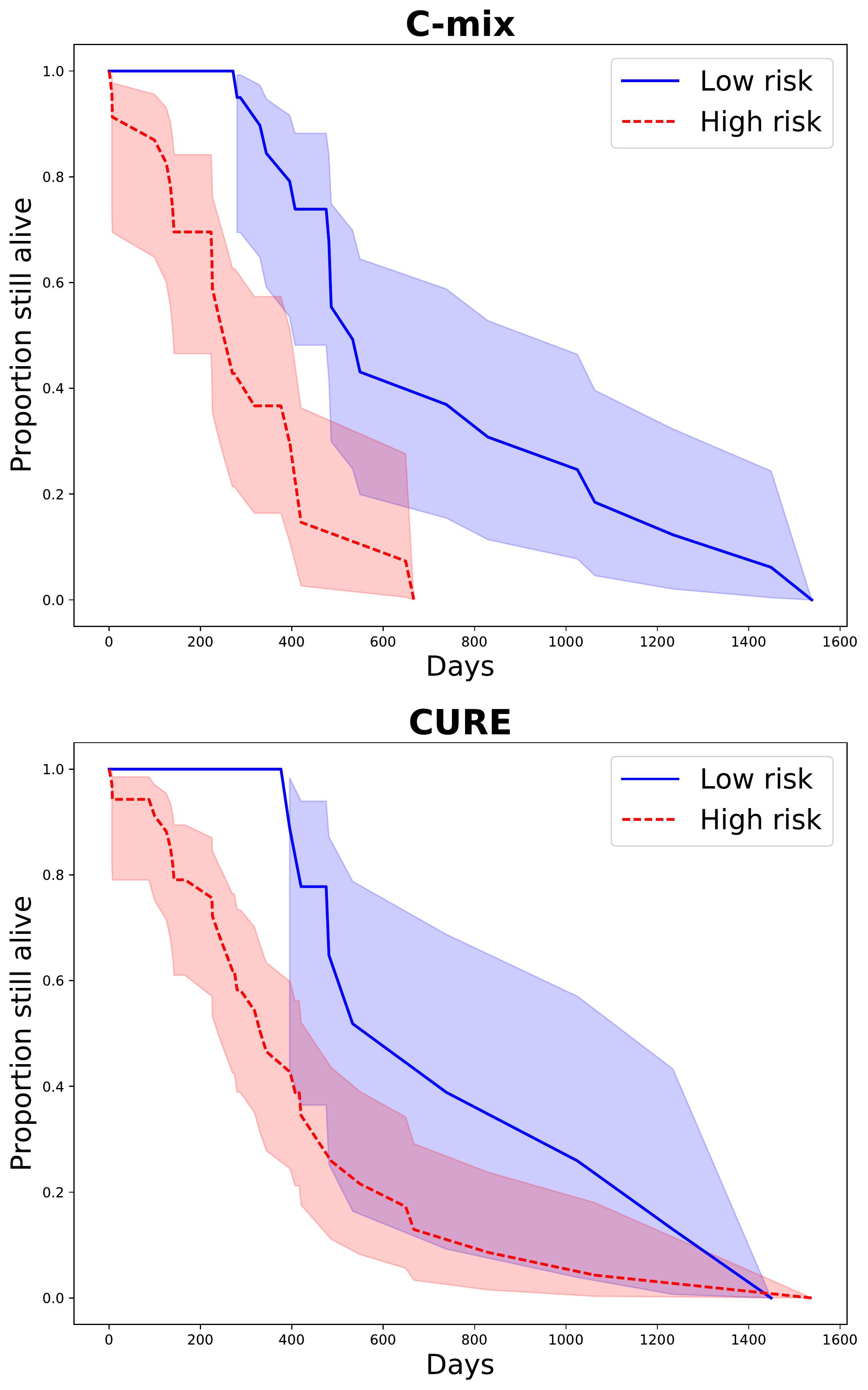}
  \caption{GBM}
  \label{fig:sub2}
\end{subfigure}
\begin{subfigure}{.33\textwidth}
  \centering
  \includegraphics[width=1\linewidth]{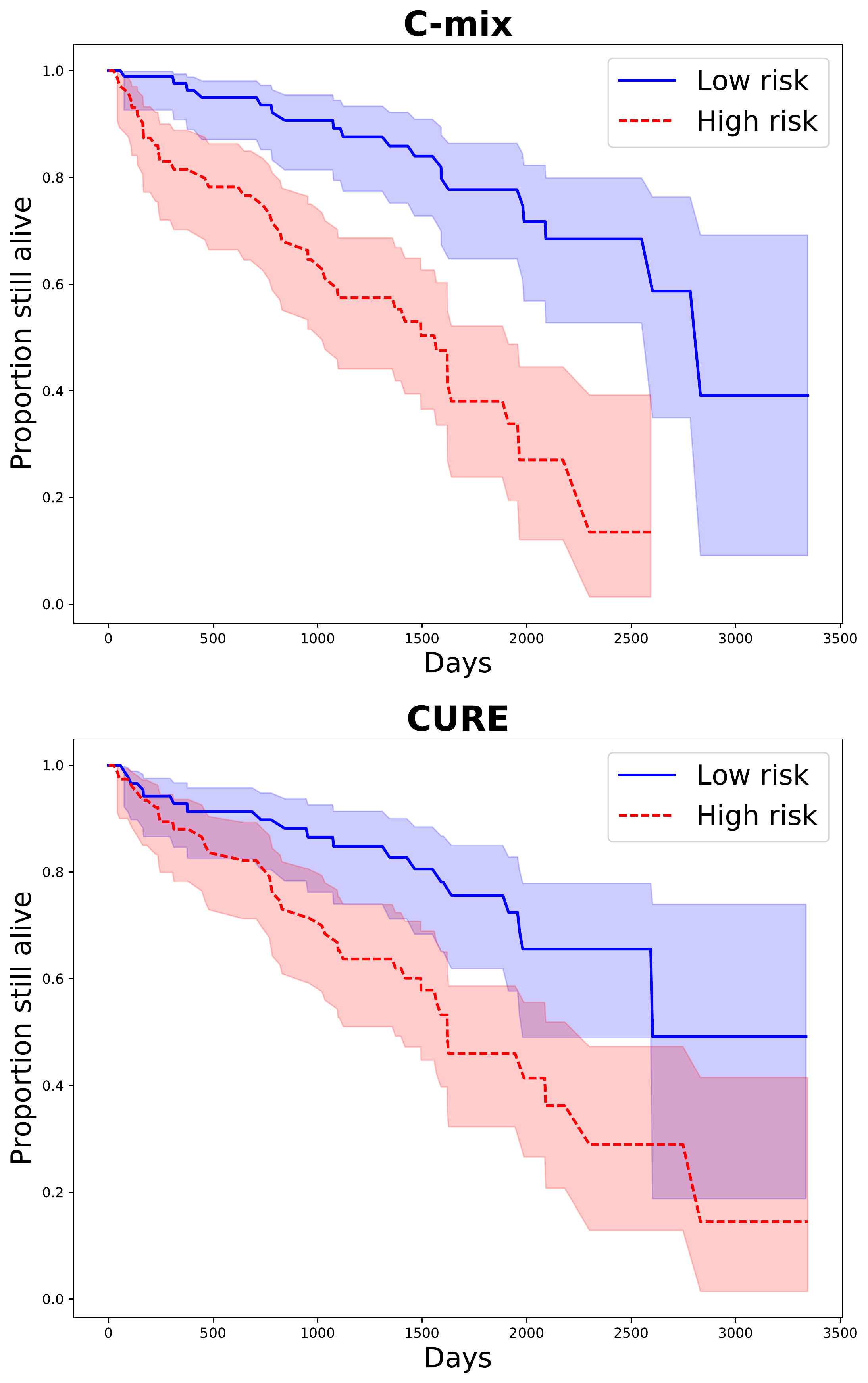}
  \caption{KIRC}
  \label{fig:sub2}
\end{subfigure}
\captionsetup{justification=justified}
\caption{Estimated survival curves per subgroups (blue for low risk and red for high risk) with the corresponding 95 \% confidence bands for the C-mix and CURE models: BRCA in column (a), GBM in column (b) and KIRC in column (c).}
\label{fig:survival-curves}
\end{figure}

A more direct approach to compare performances between models, rather than only focus on the marker order aspect, is to predict the survival of patients in the test set within a specified short time.
For the Cox PH model, the survival $\bP[T_i > t|X_i=x_i]$ for patient $i$ in the test set is estimated by
\begin{equation*}
\hat S_i(t|X_i=x_i) = [\hat S_0^\text{cox}(t)]^{\exp(x_i^\top \hat{\beta}^{\text{cox}})},
\end{equation*}
where $\hat S_0^\text{cox}$ is the estimated survival function of baseline population ($x = 0$) obtained  using the Breslow estimate of $\lambda_0$ \citep{breslow1972contribution}.
For the CURE or the C-mix models, it is naturally estimated by 
\begin{equation*}
\hat S_i(t|X_i=x_i) = \pi_{\hat{\beta}}(x_i)\hat S_1(t) + \big(1 - \pi_{\hat{\beta}}(x_i) \big) \hat S_0(t),
\end{equation*} 
where $\hat S_0$ and $\hat S_1$ are the Kaplan-Meier estimators \citep{kaplan1958nonparametric} of the low and high risk subgroups respectively, learned by the C-mix or CURE models (patients with $\pi_{\hat{\beta}}(x_i) > 0.5$ are clustered in the high risk subgroup, others in the low risk one). The corresponding estimated survival curves are given in Figure~\ref{fig:survival-curves}.
We observe that the subgroups obtained by the C-mix are more clearly separated in terms of survival than those obtained by the CURE model.

\begin{figure}[!htb]
\centering
\captionsetup[subfigure]{justification=centering}
\begin{subfigure}{.33\textwidth}
  \centering
  \includegraphics[width=1\linewidth]{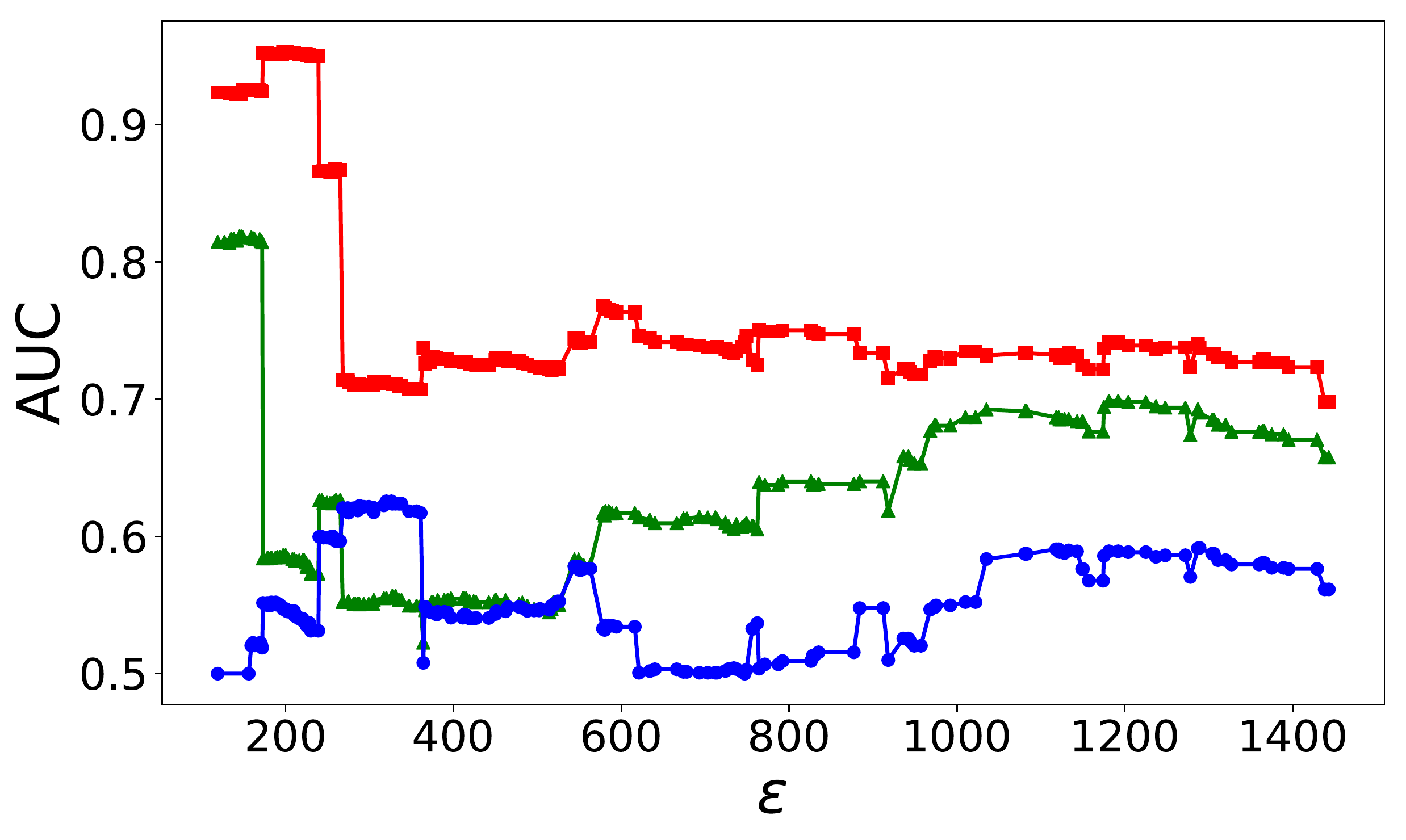}
  \caption{BRCA}
  \label{fig:sub1}
\end{subfigure}%
\begin{subfigure}{.33\textwidth}
  \centering
  \includegraphics[width=1\linewidth]{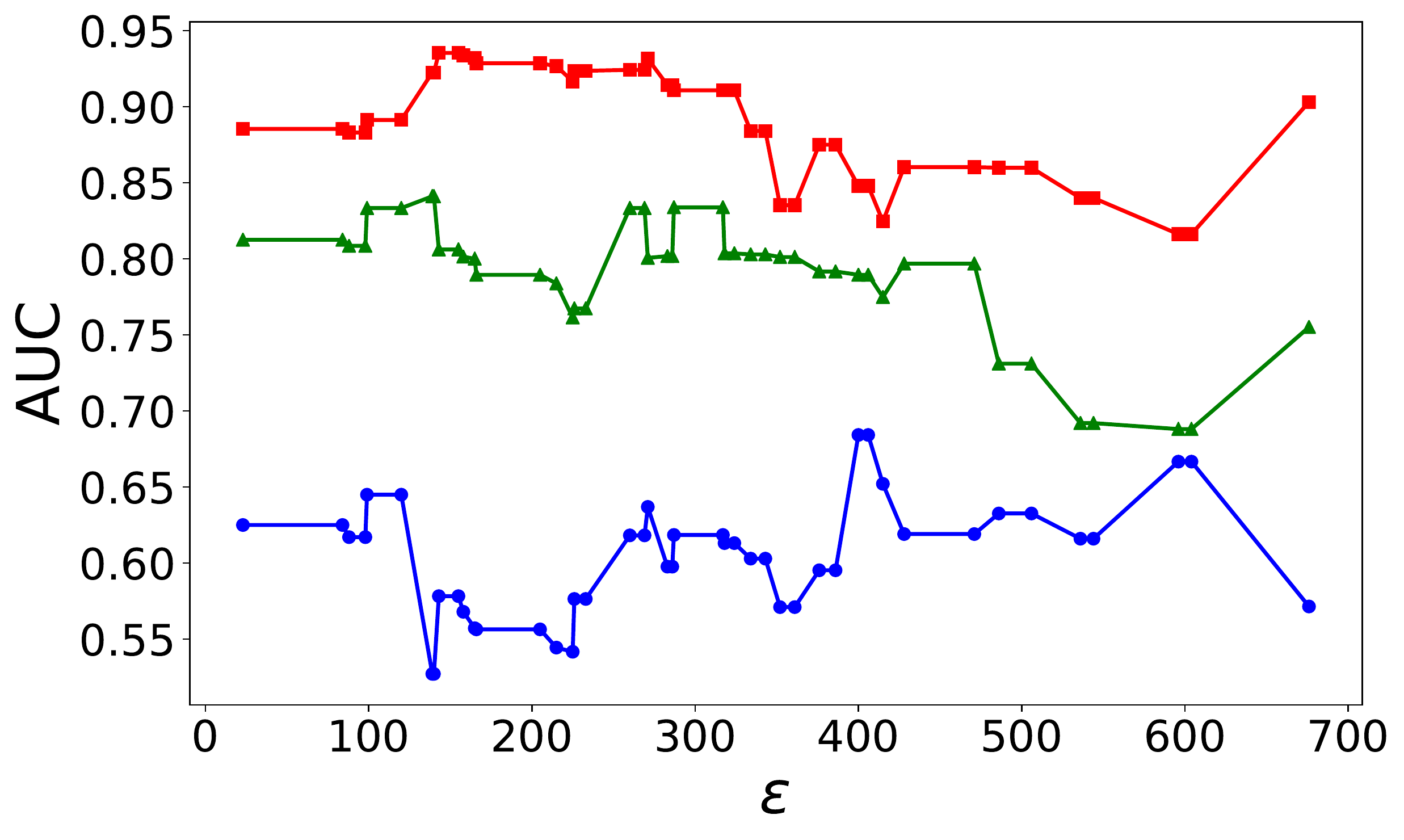}
  \caption{GBM}
  \label{fig:sub2}
\end{subfigure}
\begin{subfigure}{.33\textwidth}
  \centering
  \includegraphics[width=1\linewidth]{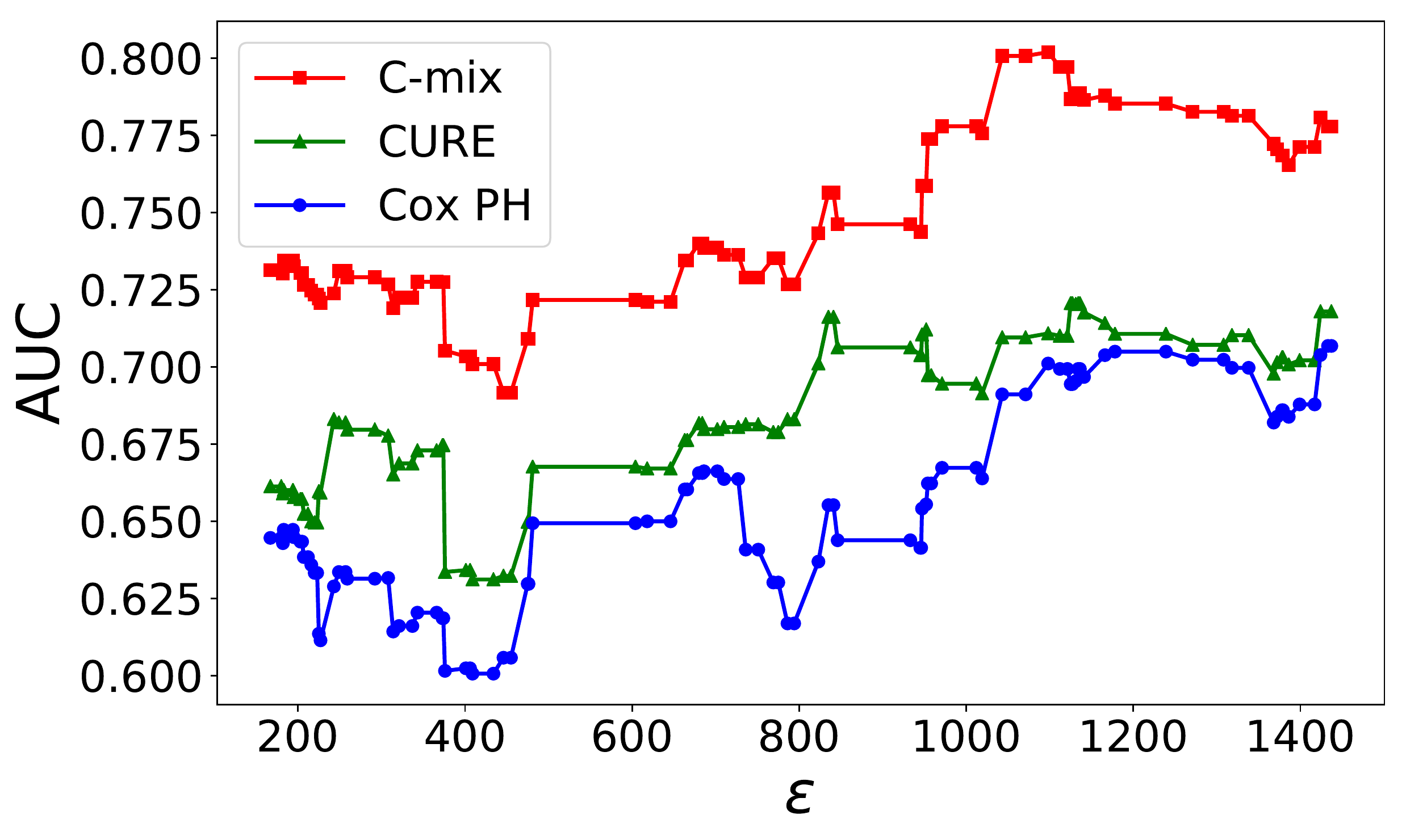}
  \caption{KIRC}
  \label{fig:sub2}
\end{subfigure}
\captionsetup{justification=justified}
\caption{Comparison of the survival prediction performances between models on the three TCGA data sets considered (still with $d=300$). Performances are, onces again, much better for the C-mix over the two other standard methods.}
\label{fig:AUC-survival}
\end{figure}

For a given time $\epsilon$, one can now use $\hat S_i(\epsilon|X_i=x_i)$ for each model to predict whether or not $T_i > \epsilon$ on the test set, resulting on a binary classification problem that we assess using the classical AUC score. By moving $\epsilon$ within the first years of follow-up, since it is the more interesting for physicians in practice, one obtains the curves given in Figure~\ref{fig:AUC-survival}.

Let us now focus on the runtime comparison between the models in Table~\ref{table:Computing time}. We choose the BRCA dataset to illustrate this point, since it is the larger one ($n=1211$) and consequently provides more clearer time-consuming differences.
\begin{table}[!htb]
\centering
\captionsetup{justification=justified}
\caption{Computing time comparison in second on the BRCA dataset ($n=1211$), with corresponding C-index in parenthesis and best result in bold in each case. This times concern the learning task for each model with the best hyper parameter selected after the cross validation procedure.
It turns out that our method is by far the fastest in addition to providing the best performances. In particular, the QNEM algorithm is faster than the \texttt{R} implementation \texttt{glmnet}.}
\begin{tabular}{ccccc}
\toprule
Model & & C-mix & CURE & Cox PH \\
\midrule
 & 100 & \textbf{0.025} (\textbf{0.792}) & 1.992 (0.764) & 0.446 (0.705) \\
$d$ & 300 & \textbf{0.027} (\textbf{0.782}) & 2.343 (0.753) & 0.810 (0.723) \\
 & 1000 & \textbf{0.139} (\textbf{0.817}) & 12.067 (0.613) & 2.145 (0.577) \\
\bottomrule
\end{tabular}
\label{table:Computing time}
\end{table}

\begin{figure}[!htb]
\begin{center}
\centerline{\includegraphics[width=10cm]{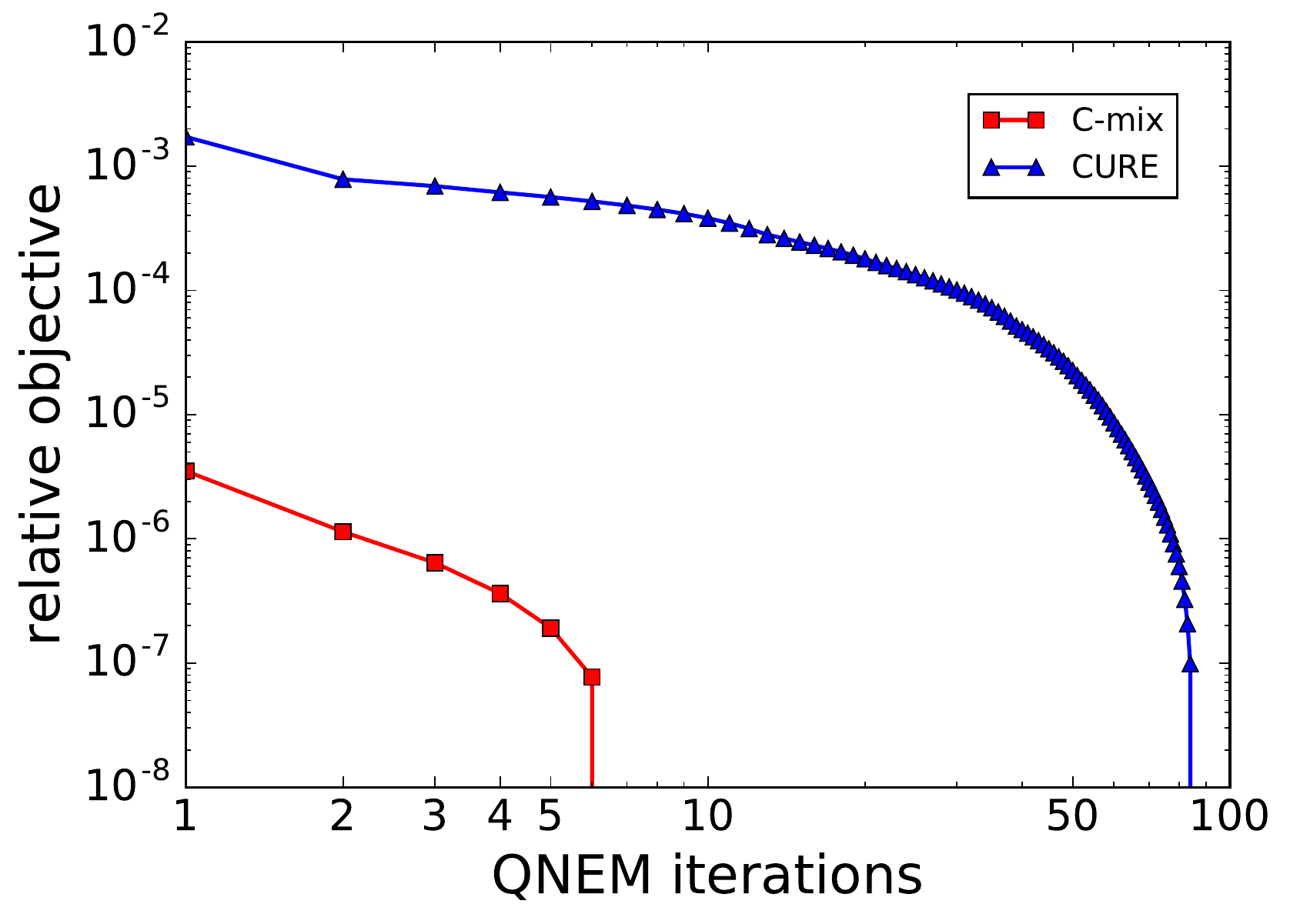}}
\end{center}
\captionsetup{justification=justified}
\caption{Convergence comparison between C-mix and CURE models through the QNEM algorithm. The relative objective is here defined at iteration $l$ as $\big( \ell_n^{\text{pen}}(\theta^{(l)}) - \ell_n^{\text{pen}}(\hat\theta)\big) / \ell_n^{\text{pen}}(\hat\theta)$, where $\hat\theta$ is naturally the parameter vector returned at the end of the QNEM algorithm, that is once convergence is reached. Note that both iteration and relative objective axis are $log$-scaled for clarity. We observe that convergence for the C-mix model is dramaticaly faster than the CURE one.}
\label{fig:cvg comparison}
\end{figure}

\noindent We also notice that despite using the same QNEM algorithm steps, our CURE model implementation is slower since convergence takes more time to be reached, as shows Figure~\ref{fig:cvg comparison}.

In Section~\ref{sec:selected genes} of Appendices, the top 20 selected genes for each cancer type and for all models are presented (for $d=300$). Literature on those genes is mined to estimate two simple scores that provide information about how related they are to cancer in general first, and second to cancer plus the survival aspect, according to scientific publications.
It turns out that some genes have been widely studied in the literature (e.g. FLT3 for the GBM cancer), while for others, very few publications were retrieved (e.g. TRMT2B still for the GBM cancer). 

\section{Concluding remarks \label{sec:concluding remarks}}
In this paper, a mixture model for censored durations (C-mix) has been introduced, and a new efficient estimation algorithm (QNEM) has been derived, that considers a penalization of the likelihood in order to perform covariate selection and to prevent overfitting.
A strong improvement is provided over the CURE and Cox PH approches (both penalized by the Elastic-Net), which are, by far, the most widely used for biomedical data analysis.
But more importantly, our method detects relevant subgroups of patients regarding their risk in a supervised learning procedure, and takes advantage of this identification to improve survival prediction over more standard methods.
An extensive Monte Carlo simulation study has been carried out to evaluate the performance of the developed estimation procedure. It showed that our approach is robust to model misspecification.
The proposed methodology has then been applied on three high dimensional datasets.
On these datasets, C-mix outperforms both Cox PH and CURE, in terms of AUC($t$), C-index or survival prediction. Moreover, many gene expressions pinpointed by the feature selection aspect of our regularized method are relevant for medical interpretations (e.g. NFKBIA, LEF1, SUSD3 or FAIM3 for the BRCA cancer, see~\citet{zhou2007enhanced} or~\citet{oskarsson2011breast}), whilst others must involve further investigations in the genetic research community.
Finally, our analysis provides, as a by-product, a new robust implementation of CURE models in high dimension.

\section*{Software}
All the methodology discussed in this paper is implemented in \texttt{Python}. The code is available from \href{https://github.com/SimonBussy/C-mix}%
{https://github.com/SimonBussy/C-mix} in the form of annotated programs, together with a notebook tutorial.

\section*{Acknowledgements}
The results shown in this paper are based upon data generated by the TCGA Research Network and freely available from \href{http://cancergenome.nih.gov/}%
{http://cancergenome.nih.gov/}. \textit{Conflict of Interest}: None declared.

\appendix

\begin{center}
\LARGE \textbf{Appendices}
\end{center}

\section{Numerical details\label{sec:Numerical details}}
Let us first give some details about the starting point of Algorithm 1. For all $k \in \{ 0, \ldots, K-1 \}$, we simply use $\beta_k^{(0)}$ as the zero vector, and for $\alpha_k^{(0)}$ we fit a censored parametric mixture model on $(y_i)_{i=1, \dots , n}$ with an EM algorithm.

Concerning the V-fold cross validation procedure for tuning $\gamma_k$, we use $V = 5$ and the cross-validation metric is the C-index. 
Let us precise that we choose $\gamma_k$ as the largest value such that error is within one standard error of the minimum, and that a grid-search is made during the cross-validation on an interval $[\gamma_k^{\text{max}} \times 10^{-4}, \gamma_k^{\text{max}} ]$, with $\gamma_k^{\text{max}}$ the interval upper bound computed in the following.

Let us consider the following convex minimization problem resulting from Equation (8), at a given step $l$:
\begin{equation*}
\hat \beta_k \in \underset{\beta \in \R^d}{\text{argmin}}\  R_{n, k}^{(l)}(\beta) + \gamma_k \big( (1-\eta)\norm{\beta}_1 + \frac{\eta}{2} \norm{\beta}_2^2 \big).
\end{equation*}
Regarding the grid of candidate values for $\gamma_k$, we consider $\gamma_k^1 \leq \gamma_k^2 \leq \dots \leq \gamma_k^{\text{max}}$. At $\gamma_k^{\text{max}}$, all coefficients $\hat \beta_{k, j}$ for $j \in \{1, \dots , d \}$ are exactly zero.
The KKT conditions \citep{boyd2004convex} claim that
\begin{center}
  $\left\{
      \begin{aligned}
        &\frac{\partial R_{n, k}^{(l)}(\hat \beta_k)}{\partial \beta_j} =  \gamma_k (1-\eta)\ \text{sgn}(\hat \beta_{k,j}) + \eta \hat \beta_{k,j}  \quad \ \ \ \forall j \in \hat \cA_k \\
        &\left| \frac{\partial R_{n, k}^{(l)}(\hat \beta_k)}{\partial \beta_j} \right| < \gamma_k (1 - \eta) \quad \quad \quad \quad \quad \quad \quad \ \forall j \notin \hat \cA_k
      \end{aligned}
    \right.$,
\end{center}
where $\hat \cA_k = \big\{ j \in \{1, \dots , d \} : \hat \beta_{k,j} \neq 0 \big\}$ is the active set of the $\hat \beta_k $ estimator, and for all $x \in~\R\setminus~\{0\},\ \text{sgn}(x) = \mathds{1}_{\{x>0\}} - \mathds{1}_{\{x<0\}} $. Then, using (10), one obtains
\begin{align*}
\forall j \in \{1, \dots , d \},\ \hat \beta_{k, j} = 0 &\Rightarrow \forall j \in \{1, \dots , d \},\ \left| n^{-1} \sum_{i=1}^n q_{i, k}^{(l)}\ \frac 12\ x_{ij}  \right| < \gamma_k (1 - \eta)
\end{align*}
Hence, we choose the following upper bound for the grid search interval during the cross-validation procedure
\begin{equation*}
\gamma_k^{\text{max}} = \frac{1}{2n(1-\eta)} \underset{j \in \{1, \dots , d \}}{\text{max}} \sum_{i=1}^n |x_{ij}|.
\end{equation*}

\section{Proof of Theorem 1 \label{sec:th1 proof}} 

Let us denote $D = \sum_{k=0}^{K-1} d_k + Kd$ the number of coordinates of $\theta$ so that one can write 
\begin{equation*}
\theta =(\theta_1, \dots, \theta_D) = (\alpha_0, \ldots, \alpha_{K-1}, \beta_0, \ldots, \beta_{K-1})^\top \in \Theta \subset \R^D.
\end{equation*}
We denote $\bar{\theta}$ a cluster point of the sequence $S = \{ \theta^{(l)}; l=0,1,2,\dots \}$ generated by the QNEM algorithm, $i.e.\ \forall \varepsilon > 0, V_\varepsilon(\bar{\theta}) \cap S\setminus \{\bar{\theta}\} \neq \varnothing$, with $V_\varepsilon(\bar{\theta})$ the epsilon-neighbourhood of $\bar{\theta}$. We want to prove that $\bar{\theta}$ is a stationary point of the non-differentiable function $\theta \mapsto \ell^{\text{pen}}_n(\theta)$, which means \citep{tseng2001convergence}:
\begin{equation}
  \label{eq:directionalderivative}
\forall r \in \R^{D}, \nu^{pen\ '}_n(\bar{\theta} ; r)=\underset{\zeta \rightarrow 0}{\text{lim}} \frac{\ell^{\text{pen}}_n(\bar{\theta} + r \zeta) - \ell^{\text{pen}}_n(\bar{\theta})}{\zeta} \geq 0.
\end{equation}
The proof is inspired by~\citet{bertsekas1995nonlinear}. The conditional density of the complete data given the observed data can be written 
\begin{equation*}
k(\theta) =  \frac{\exp \big ( \ell_n^{\text{comp}}(\theta) \big )}{\exp \big ( \ell_n (\theta) \big )}.
\end{equation*}
Then, one has
\begin{equation}
  \label{eq:objfunc}
\ell^{\text{pen}}_n(\theta) = Q_n^{\text{pen}}(\theta, \theta^{(l)}) - H(\theta, \theta^{(l)}),
\end{equation}
where we introduced $H(\theta, \theta^{(l)}) = \E_{\theta^{(l)}}[\log \big( k(\theta) \big)]$. 
The key argument relies on the following facts that hold under Hypothesis (3) and (4):
\begin{itemize}
  \item $Q_n^{\text{pen}}(\theta, \theta^{(l)})$ is continuous in $\theta$ and $\theta^{(l)}$,
  \item  for any fixed $\theta^{(l)}$ (at the $(l+1)$-th M step of the algorithm), $Q_{n,\theta^{(l)}}^{\text{pen}}(\theta)$ is convex in $\theta$ and strictly convex in each coordinate of $\theta$.
\end{itemize}
Let $r \in \R^{D}$ be an arbitrary direction, then Equations~\eqref{eq:directionalderivative} and~\eqref{eq:objfunc} yield
\begin{equation*}
\ell^{\text{pen}\ '}_n(\bar{\theta} ; r) = Q_{n, \bar{\theta}}^{\text{pen}\ '}(\bar{\theta}; r) - \langle \bigtriangledown H_{\bar{\theta}}(\bar{\theta}), r \rangle.
\end{equation*}
Hence, by Jensen's inequality we get 
\begin{equation}
  \label{eq:ineqH}
\forall \theta \in \Theta, H(\theta^{(l)}, \theta^{(l)}) \leq H(\theta, \theta^{(l)}),
\end{equation}
and so $\theta \mapsto H_{\bar{\theta}}(\theta)$ is minimized for $\theta= \theta^{(l)}$, then we have $\bigtriangledown H_{\bar{\theta}}(\bar{\theta})=0$. It remains to prove that $Q_{n, \bar{\theta}}^{\text{pen}\ '}(\bar{\theta}; r) \geq 0$. Let us focus on the proof of the following expression
\begin{equation}
\label{eq:firstcoord}
\forall x_1, Q_{n, \bar{\theta}}^{\text{pen}}(\bar{\theta}) \leq Q_{n, \bar{\theta}}^{\text{pen}}(x_1,\bar{\theta}_2,\dots,\bar{\theta}_D).
\end{equation}
Denoting $w_i^{(l)}=(\theta_1^{(l+1)},\dots,\theta_i^{(l+1)},\theta_{i+1}^{(l)},\dots,\theta_D^{(l)})$ and from the definition of the QNEM algorithm, we first have 
\begin{equation}
\label{eq:ineg1}
Q_{n, \theta^{(l)}}^{\text{pen}}(\theta^{(l)}) \geq Q_{n, \theta^{(l)}}^{\text{pen}}(w_1^{(l)}) \geq \dots \geq Q_{n, \theta^{(l)}}^{\text{pen}}(w_{D-1}^{(l)}) \geq Q_{n, \theta^{(l)}}^{\text{pen}}(\theta^{(l+1)}),
\end{equation}
and second for all $x_1, Q_{n,\theta^{(l)}}^{\text{pen}}(w_1^{(l)}) \leq Q_{n,\theta^{(l)}}^{\text{pen}}(x_1,\theta_2^{(l)},\dots,\theta_D^{(l)})$.
Consequently, if $(w_1^{(l)})_{l \in \N}$ converges to $\bar{\theta}$, one obtains~\eqref{eq:firstcoord} by continuity taking the limit $l \rightarrow \infty$. Let us now suppose that $(w_1^{(l)})_{l \in \N}$ does not converge to $\bar{\theta}$, so that $(w_1^{(l)} - \theta^{(l)})_{l \in \N}$ does not converge to 0. Or equivalently: there exists a subsequence $(w_1^{(l_j)} - \theta^{(l_j)})_{j \in \N}$ not converging to 0. 

Then, denoting $\psi^{(l_j)} = \norm{w_1^{(l_j)} - \theta^{(l_j)}}_2$, we may assume that there exists $\bar{\psi} > 0$ such that $\forall j \in~\N, \psi^{(l_j)} > \bar{\psi}$ by removing from the subsequence $(w_1^{(l_j)} - \theta^{(l_j)})_{j \in \N}$ any terms for which $\psi^{(l_j)} = 0$. Let $s_1^{(l_j)} = \frac{w_1^{(l_j)} - \theta^{(l_j)}}{\psi^{(l_j)}}$, so that $(s_1^{(l_j)})_{j \in \N}$ belongs to a compact set $(\norm{s_1^{(l_j)}}=1)$ and then converges to $\bar{s_1} \neq 0$. Let us fix some $\epsilon \in [0,1]$, then $0 \leq \epsilon \bar{\psi} \leq \psi^{(l_j)}$. Moreover, $\theta^{(l_j)} + \epsilon \bar{\psi} s_1^{(l_j)}$ lies on the segment joining $\theta^{(l_j)}$ and $w_1^{(l_j)}$, and consequently belongs to $\Theta$ since $\Theta$ is convex. As $Q_{n, \theta^{(l_j)}}^{\text{pen}}(.)$ is convex and $w_1^{(l_j)}$ minimizes this function over all values that differ from $\theta^{(l_j)}$ along the first coordinate, one has
\begin{align}
\label{eq:ineg2}
Q_{n, \theta^{(l_j)}}^{\text{pen}}(w_1^{(l_j)}) &= Q_{n, \theta^{(l_j)}}^{\text{pen}}(\theta^{(l_j)} + \psi^{(l_j)} s_1^{(l_j)}) \nonumber \\ 
&\leq Q_{n, \theta^{(l_j)}}^{\text{pen}}(\theta^{(l_j)} + \epsilon \bar{\psi} s_1^{(l_j)}) \nonumber \\
&\leq Q_{n, \theta^{(l_j)}}^{\text{pen}}(\theta^{(l_j)}).
\end{align}
We finally obtain
\begin{align*}
0 &\leq Q_{n, \theta^{(l_j)}}^{\text{pen}}(\theta^{(l_j)}) - Q_{n, \theta^{(l_j)}}^{\text{pen}}(\theta^{(l_j)} + \epsilon \bar{\psi} s_1^{(l_j)}) \\
& \underset{~\eqref{eq:ineg2}}{\leq} Q_{n, \theta^{(l_j)}}^{\text{pen}}(\theta^{(l_j)}) - Q_{n, \theta^{(l_j)}}^{\text{pen}}(w_1^{(l_j)}) \\
& \underset{~\eqref{eq:ineg1}}{\leq} Q_{n, \theta^{(l_j)}}^{\text{pen}}(\theta^{(l_j)}) - Q_{n, \theta^{(l_j)}}^{\text{pen}}(\theta^{(l_j+1)}) \\
& \underset{~\eqref{eq:objfunc}}{\leq} \ell^{\text{pen}}_n(\theta^{(l_j)}) - \ell^{\text{pen}}_n(\theta^{(l_j+1)}) + \underbrace{H_{\theta^{(l_j)}}(\theta^{(l_j)}) - H_{\theta^{(l_j)}}(\theta^{(l_j+1)})}_{\underset{~\eqref{eq:ineqH}}{\leq} 0 } \\
&\leq \ell^{\text{pen}}_n(\theta^{(l_j)}) - \ell^{\text{pen}}_n(\theta^{(l_j+1)}) \underset{j \rightarrow \infty}{\longrightarrow} \ell^{\text{pen}}_n(\bar{\theta}) - \ell^{\text{pen}}_n(\bar{\theta}) = 0
\end{align*}
By continuity of the function $Q_n^{\text{pen}}(x,y)$ in both $x$ and $y$ and taking the limit $j \rightarrow \infty$, we conclude that $\forall \epsilon \in [0,1],\ Q_{n, \bar{\theta}}^{\text{pen}}(\bar{\theta} + \epsilon \bar{\psi} \bar{s_1}) = Q_{n, \bar{\theta}}^{\text{pen}}(\bar{\theta})$. Since $\bar{\psi} \bar{s_1} \neq 0$, this contradicts the strict convexity of $x_1 \mapsto Q_{n,\theta^{(l)}}^{\text{pen}}(x_1,\theta_2^{(l)},\dots,\theta_D^{(l)})$ and establishes that $(w_1^{(l)})_{l \in \N}$ converges to $\bar{\theta}$.

Hence~\eqref{eq:firstcoord} is proved. Repeating the argument to each coordinate, we deduce that $\bar{\theta}$ is a coordinate-wise minimum, and finally conclude that $\ell_n^{\text{pen}\ '}(\bar{\theta}; r) \geq 0$ \citep{tseng2001convergence}. Thus, $\bar{\theta}$ is a stationary point of the criterion function defined in Equation (4).

$\hfill{\square}$
\section{Additional comparisons \label{sec:additional comparison}}
In this section, we consider two extra simulation settings. First, we consider the case $d \gg n$, which is the setting of our application on TCGA datasets. Then, we add another simulation case under the C-mix model using gamma distributions instead of geometric ones. The shared parameters in the two cases are given in Table~\ref{table:parameters choice}.
\begin{table}[!htb]
\centering
\caption{Hyper-parameters choice for simulation.}
\begin{tabular}{|c|c|c|c|c|c|c|c|c|}
\hline
$\eta$ & $n$ & $s$ & $r_{cf}$ & $\nu$ & $\rho$ & $\pi_0$ & gap & $r_c$  \\
\hline
0.1 & 250 & 50 & 0.5 & 1 & 0.5 & 0.75 & 0.1 & 0.5 \\
\hline
\end{tabular}
\label{table:parameters choice}
\end{table}
\subsection{Case $d \gg n$ \label{sec:high dim case}}
Data is here generated under the C-mix model with $(\alpha_0, \alpha_1) = (0.1, 0.5)$ and $d\in\{200, 500, 1000\}$. The 3 models are trained on a training set and risk prediction is made on a test set. We also compare the 3 models when a dimension reduction step is performed at first, using two different screening methods. The first is based on univariate Cox PH models, namely the one we used in Section 5 of the paper (in our application to genetic data), where we select here the top 100 variables. This screening method is hence referred as ``top 100'' in the following. The second is the iterative sure independence screening (ISIS) method introduced in \citet{fan2010high}, using the \texttt{R} package \texttt{SIS}~\cite{saldana2016sis}. Prediction performances are compared in terms of C-index, while variable selection performances are compared in terms of AUC using the method detailed in Section~\ref{sec:details feature selection}, and we also add two more classical scores~\citep{fan2010high} for comparison: the median $\ell_1$ and squared $\ell_2$ estimation errors, given by $\norm{\beta - \hat \beta}_1$ and $\norm{\beta - \hat \beta}_2$ respectively. Results are given in Table~\ref{table:simu-high-dim}.
\begin{table}[!htb]
\centering
\captionsetup{justification=justified}
\caption{Average performances and standard deviation (in parenthesis) on 100 simulated data for different dimension $d$ and different screening method (including no screening). For each configuration, the best result appears in bold.}
\resizebox{\textwidth}{!}{%
\begin{tabular}{ccccccc}
\toprule
$d$ & \textnormal{screening} & \textnormal{model} & \textnormal{C-index} & \textnormal{AUC} & $\norm{\beta - \hat \beta}_1$ & $\norm{\beta - \hat \beta}_2$ \\
\midrule
            &          & C-mix  & \textbf{0.716 (0.062)} & \textbf{0.653 (0.053)} &  \textbf{51.540 (0.976)} &  \textbf{7.254 (0.129)}  \\
            &  none    &  CURE  & 0.701 (0.067) & 0.625 (0.052) &  51.615 (1.275) &  7.274 (0.122)  \\
            &          & Cox PH & 0.672 (0.089) & 0.608 (0.063) & 199.321 (0.490) &  99.679 (0.229) \\
            \cmidrule(l){2-7}
            &          & C-mix  & \textbf{0.737 (0.057)} & \textbf{0.682 (0.060)} &  \textbf{52.297 (1.351)} &  \textbf{7.381 (0.161)}  \\
    200     & top 100  &  CURE  & 0.714 (0.060) & 0.651 (0.050) &  52.366 (1.382) &  7.386 (0.134)  \\
            &          & Cox PH & 0.692 (0.089) & 0.630 (0.070) &  52.747 (0.530) &  7.946 (0.093)  \\
            \cmidrule(l){2-7}
            &          & C-mix  & \textbf{0.691 (0.049)} & 0.570 (0.011) &  55.493 (1.624) &  8.083 (0.394)  \\
            &   ISIS   &  CURE  & 0.685 (0.050) & 0.571 (0.009) &  54.461 (1.112) &  7.848 (0.211)  \\
            &          & Cox PH & 0.690 (0.049) & \textbf{0.573 (0.011)} &  \textbf{48.186 (0.366)} &  \textbf{6.840 (0.037)}  \\
\midrule
            &          & C-mix  & \textbf{0.710 (0.058)} & \textbf{0.642 (0.057)} &  \textbf{51.627 (0.994)} &  7.277 (0.106)  \\
            &  none    &  CURE  & 0.675 (0.057) & 0.610 (0.052) &  51.920 (2.411) &  \textbf{7.252 (0.138)}  \\
            &          & Cox PH & 0.624 (0.097) & 0.567 (0.057) & 499.610 (0.396) & 157.997 (0.117) \\
            \cmidrule(l){2-7}
            &          & C-mix  & \textbf{0.735 (0.050)} & \textbf{0.694 (0.057)} &  53.161 (1.708) &  7.433 (0.152)  \\
    500     & top 100  &  CURE  & 0.703 (0.054) & 0.649 (0.042) &  53.419 (1.818) &  7.387 (0.133)  \\
            &          & Cox PH & 0.682 (0.087) & 0.633 (0.074) &  \textbf{49.465 (0.428)} &  \textbf{6.937 (0.094)}  \\
            \cmidrule(l){2-7}
            &          & C-mix  & \textbf{0.677 (0.051)} & 0.559 (0.013) &  55.229 (1.831) &  7.974 (0.375)  \\
            &   ISIS   &  CURE  & 0.671 (0.051) & 0.559 (0.015) &  54.187 (1.244) &  7.754 (0.227)  \\
            &          & Cox PH & 0.675 (0.051) & \textbf{0.560 (0.016)} &  \textbf{48.574 (0.614)} &  \textbf{6.870 (0.054)}  \\
\midrule
            &          & C-mix  & \textbf{0.694 (0.063)} & \textbf{0.633 (0.066)} &  \textbf{51.976 (1.921)} &  \textbf{7.272 (0.141)}  \\
            &  none    &  CURE  & 0.657 (0.067) & 0.598 (0.057) &  52.078 (2.414) &  7.236 (0.138)  \\
            &          & Cox PH & 0.579 (0.092) & 0.541 (0.050) & 999.768 (0.316) & 223.558 (0.067) \\
            \cmidrule(l){2-7}
            &          & C-mix  & \textbf{0.726 (0.050)} & \textbf{0.693 (0.040)} &  53.813 (1.592) &  7.149 (0.115)  \\
    1000    & top 100  &  CURE  & 0.685 (0.061) & 0.653 (0.037) &  54.146 (1.596) &  7.383 (0.090)  \\
            &          & Cox PH & 0.688 (0.076) & 0.668 (0.064) &  \textbf{52.838 (0.558)} &  \textbf{6.909 (0.077)}  \\
            \cmidrule(l){2-7}
            &          & C-mix  & \textbf{0.653 (0.062)} & 0.553 (0.017) &  53.760 (1.949) &  7.269 (0.395)  \\
            &   ISIS   &  CURE  & 0.652 (0.061) & \textbf{0.554 (0.015)} &  53.928 (1.288) &  7.687 (0.236)  \\
            &          & Cox PH & 0.652 (0.063) & 0.553 (0.015) &  \textbf{51.826 (0.606)} &  \textbf{6.895 (0.054)}  \\
\bottomrule
\end{tabular}}
\label{table:simu-high-dim}
\end{table}

The C-mix model obtains constantly the best C-index performances in prediction, for all settings. Moreover, the ``top 100'' screening method improve the 3 models prediction power, while ISIS method only improve the Cox PH model prediction power. As expected, ISIS method significantly improve the Cox PH model in terms of variable selection and obtains the best results for $d=500$ and 1000. Conclusions in terms of variable selection are the same relatively to the AUC, $\ell_1$ and squared $\ell_2$ estimation errors. Then, in the paper, we only focus on the AUC method detailed in Section~\ref{sec:details feature selection}. Note that the Cox PH model obtains the best results in terms of variable selection with the two screening method, since both screening methods are based on the Cox PH model. Thus, one could improve the C-mix variable selection performances by simply use the ``top 100'' screening method with univariate C-mix, which was not the purpose of the section. Finally, the results obtained justify the screening strategy we use in Section 5 of the paper.

\subsection{Case of times simulated with a mixture of gammas}
We consider here the case where data is simulated under the C-mix model with gamma distributions instead of geometric ones, not to confer to the C-mix prior information on the underlying survival distributions. Hence, one has 
\begin{equation*}
f_k(t;\iota_k, \zeta_k) = \frac{t^{\iota_k - 1} e^{- \frac{t}{\zeta_k}}}{\zeta_k^{\iota_k}\Gamma(\iota_k)},
\end{equation*}
with $\iota_k$ the shape parameter, $\zeta_k$ the scale parameter and $\Gamma$ the gamma function. For the simulations, we choose $(\iota_0, \zeta_0) = (5, 3)$ and $(\iota_1, \zeta_1) = (1.5, 1)$, so that density and survival curves are comparable with those in Section~\ref{sec:high dim case}, as illustrates Figure~\ref{fig:distributions-comparison} below.
\begin{figure}[!htb]
\centering
\captionsetup[subfigure]{justification=centering}
\begin{subfigure}{.49\textwidth}
  \centering
  \includegraphics[width=1\linewidth]{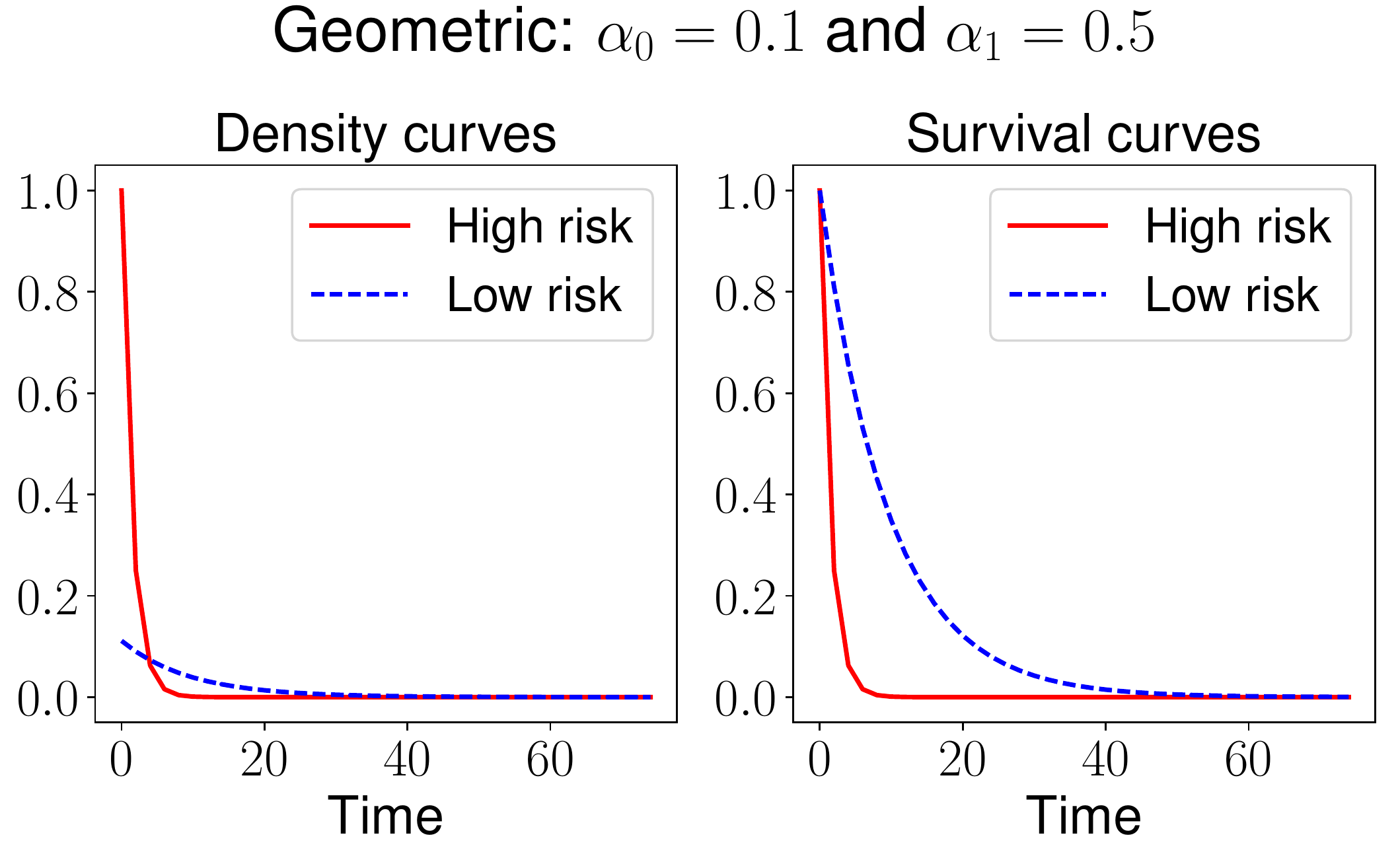}
\end{subfigure}
\begin{subfigure}{.49\textwidth}
  \centering
  \includegraphics[width=1\linewidth]{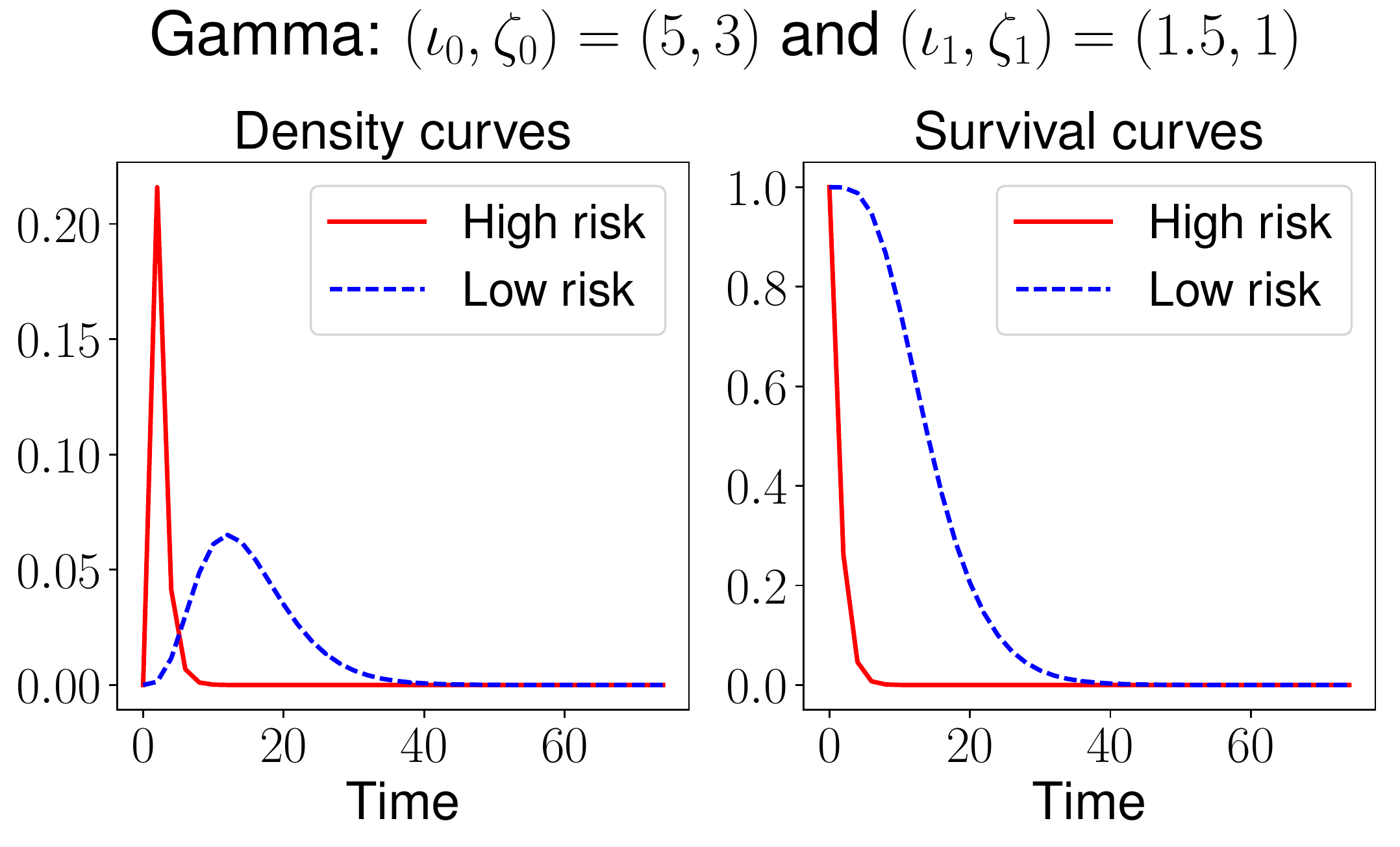}
\end{subfigure}
\captionsetup{justification=justified}
\caption{Comparison of the density and survival curves of geometrics laws used in Section~\ref{sec:high dim case} and those used in this section. The supports are then relatively close.}
\label{fig:distributions-comparison}
\end{figure}

  We also add another class of model for comparison in this context: the accelerated failure time model~\citep{wei1992accelerated} (AFT); which can be viewed as a parametric Cox model. Indeed, the semi-parametric property of the Cox PH model could lower its performances compared to completely parametric models such as C-mix and CURE ones, especially in simulations where $n$ is relatively small. We use the \texttt{R} package \texttt{AdapEnetClass} that implements AFT in a high dimensional setting using two Elastic-Net regularization approaches~\citep{khan2016variable}: the adaptive Elastic-Net (denoted AEnet in the following) and the weighted Elastic-Net (denoted WEnet in the following). Results are given in Table~\ref{table:simu-gamma} using the same metrics that in Section~\ref{sec:high dim case}.

\begin{table}[!htb]
\centering
\captionsetup{justification=justified}
\caption{Average performances and standard deviation (in parenthesis) on 100 simulated data for different dimension $d$ with the times simuted with a mixture of gammas. For each configuration, the best result appears in bold.}
\resizebox{\textwidth}{!}{%
\begin{tabular}{cccccc}
\toprule
$d$ & \textnormal{model} & \textnormal{C-index} & \textnormal{AUC} & $\norm{\beta - \hat \beta}_1$ & $\norm{\beta - \hat \beta}_2$ \\
\midrule
            & C-mix  & \textbf{0.701 (0.090)} & \textbf{0.659 (0.083)} &  \textbf{51.339 (2.497)} &  \textbf{7.186 (0.281)}  \\
            &  CURE  & 0.682 (0.058) & 0.609 (0.037) &  51.563 (1.071) &  7.263 (0.097)  \\
    200     & Cox PH & 0.664 (0.085) & 0.605 (0.065) & 199.337 (0.493) &  99.686 (0.231) \\
            & AEnet  & 0.631 (0.062) & 0.577 (0.046) &  54.651 (2.328) &  7.713 (0.426)  \\
            & WEnet  & 0.620 (0.061) & 0.544 (0.030) &  58.861 (4.298) &  8.568 (0.851)  \\
    \midrule
            & C-mix  & \textbf{0.704 (0.100)} & \textbf{0.651 (0.084)} &  52.416 (2.311) &  7.357 (0.231)  \\
            &  CURE  & 0.687 (0.057) & 0.609 (0.038) &  \textbf{52.041 (1.667)} &  \textbf{7.262 (0.096)}  \\
    500     & Cox PH & 0.621 (0.101) & 0.559 (0.057) & 499.677 (0.381) & 158.017 (0.113) \\
            & AEnet  & 0.604 (0.061) & 0.557 (0.030) &  55.126 (1.693) &  7.616 (0.316)  \\
            & WEnet  & 0.594 (0.065) & 0.535 (0.021) &  59.736 (2.777) &  8.438 (0.626)  \\
    \midrule
            & C-mix  & \textbf{0.684 (0.097)} & \textbf{0.638 (0.088)} &  \textbf{52.557 (3.746)} &  7.331 (0.277)  \\
            &  CURE  & 0.658 (0.057) & 0.603 (0.044) &  53.120 (3.853) &  \textbf{7.273 (0.165)}  \\
    1000    & Cox PH & 0.580 (0.092) & 0.538 (0.053) & 999.785 (0.334) & 223.561 (0.071) \\
            & AEnet  & 0.586 (0.058) & 0.541 (0.024) &  54.597 (1.312) &  7.495 (0.299)  \\
            & WEnet  & 0.583 (0.054) & 0.525 (0.017) &  58.746 (2.260) &  8.150 (0.551)  \\
\bottomrule
\end{tabular}}
\label{table:simu-gamma}
\end{table}

Hence, the C-mix model still gets the best results, both in terms of risk prediction and variable selection. Note that AFT with AEnet and WEnet outperforms the Cox model regularized by the Elastic-Net when $d=1000$, but is still far behind the C-mix performances.

\section{Tuning of the censoring level \label{sec:censoring level}}
Suppose that we want to generate data following the procedure detailed in Section 4.2, in the C-mix with geometric distributions or CURE case. 
The question here is to choose $\alpha_c$ for a desired censoring rate $r_c$, and for some fixed parameters $\alpha_0$, $\alpha_1$ and $\pi_0$. 
We write
\begin{align*}
  1 - r_c &= \E[\delta] = \sum_{k=0}^{+\infty} \sum_{j=1}^{+\infty} \big[ \alpha_0(1-\alpha_0)^{j-1}\pi_0 + \alpha_1(1-\alpha_1)^{j-1}(1-\pi_0) \big]\alpha_c(1-\alpha_c)^{j+k-1} \\
  &= \frac{\alpha_0\pi_0\big[1-(1-\alpha_1)(1-\alpha_c)\big] + \alpha_1(1-\pi_0)\big[1-(1-\alpha_0)(1-\alpha_c)\big]}{\big[1-(1-\alpha_0)(1-\alpha_c)\big]\big[1-(1-\alpha_1)(1-\alpha_c)\big]}.
\end{align*}
Then, if we denote $\bar{r}_c=1-r_c$, $\bar{\alpha}_c=1-\alpha_c$, $\bar{\alpha}_0=1-\alpha_0$, $\bar{\alpha}_1=1-\alpha_1$ and $\bar{\pi}_0=1-\pi_0$, we can choose $\alpha_c$ for a fixed $r_c$ by solving the following quadratic equation
\begin{equation*}
(\bar{r}_c\ \bar{\alpha}_0\ \bar{\alpha}_1)\bar{\alpha}_c^2 + \big(\alpha_0\pi_0\bar{\alpha}_1+\alpha_1\bar{\pi}_0\ \bar{\alpha}_0-\bar{r}_c(\bar{\alpha}_1+\bar{\alpha}_0)\big)\bar{\alpha}_c+(r_c-\alpha_0\pi_0-\alpha_1\bar{\pi}_0)=0  ,
\end{equation*}
for which one can prove that there is always a unique root in $(0, 1)$.

\section{Details on variable selection evaluation \label{sec:details feature selection}}
\begin{figure}[!htb]
\centering
\captionsetup[subfigure]{justification=centering}
\begin{subfigure}{\textwidth}
  \centering
  \includegraphics[width=\linewidth]{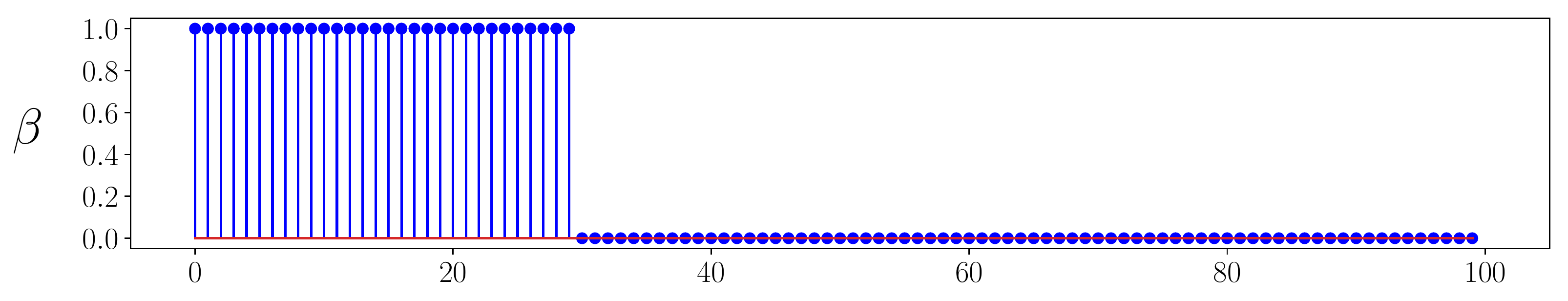}
\end{subfigure}
\begin{subfigure}{\textwidth}
  \centering
  \includegraphics[width=1\linewidth]{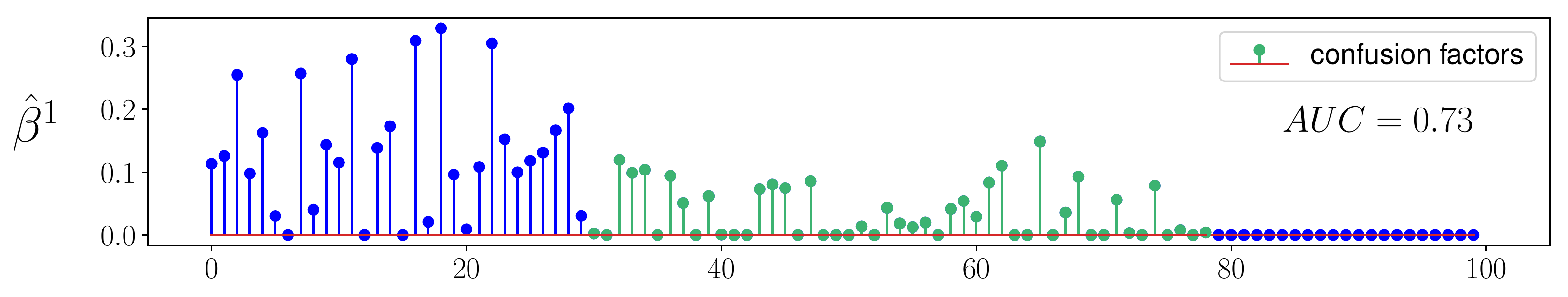}
\end{subfigure}
\begin{subfigure}{\textwidth}
  \centering
  \includegraphics[width=1\linewidth]{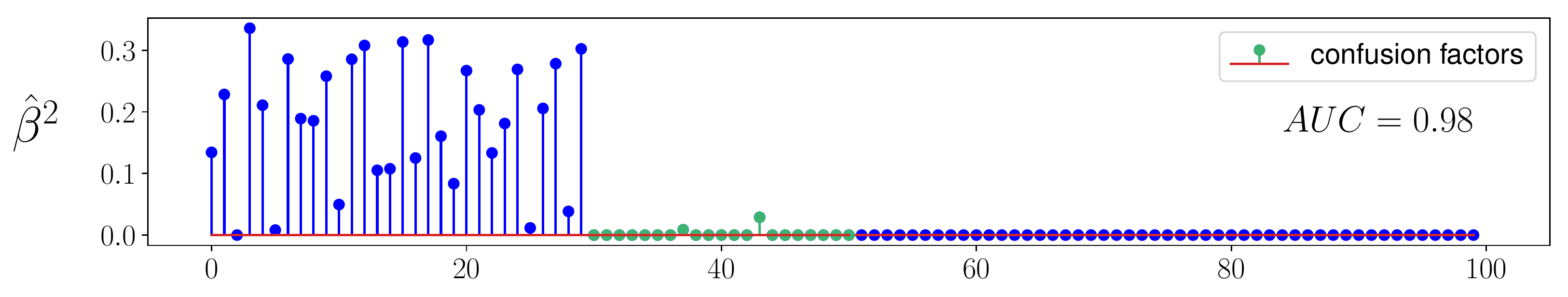}
\end{subfigure}
\captionsetup{justification=justified}
\caption{Illustration of the variable selection evaluation procedure. $\hat \beta^1$ is learned by the C-mix according to data generated with $\beta$ and $(\text{gap}, r_{cf}) = (0.2, 0.7)$. We observe that using this gap value to generate data, the model does not succeed to completely vanish the confusion variables (being $70\%$ of the non-active variables, represented in green color), while all other non-active variables are vanished. The corresponding AUC scrore of feature selection is 0.73. $\hat \beta^2$ is learned by the C-mix according to data generated with $\beta$ and $(\text{gap}, r_{cf}) = (1, 0.3)$. The confusion variables are here almost all detected and the corresponding AUC scrore of feature selection is 0.98.}
\label{fig:variable-selection-evaluation}
\end{figure}

Let us recall that the true underlying $\beta$ used in the simulations is given by 
\begin{equation*}
\beta=~(\underbrace{\nu,\ldots,\nu}_s,0,\ldots,0) \in \R^d,
\end{equation*}
with $s$ the sparsity parameter, being the number of ``active'' variables. To illustrate how we assess the variable selection ability of the considered models, we give in Figure~\ref{fig:variable-selection-evaluation} an example of $\beta$ with $d=100$, $\nu=1$ and $s=30$. We simulate data according to this vector (and to the C-mix model) with two different $(\text{gap}, r_{cf})$ values: $(0.2, 0.7)$ and $(1, 0.3)$. Then, we give the two corresponding estimated vectors $\hat \beta$ learned by the C-mix on this data.

Denoting $\tilde{\beta}_i = |\hat{\beta}_i| / \text{max} \big\{ |\hat{\beta}_i|, i \in \{1,\ldots,d \} \big\}$, we consider that $\tilde{\beta}_i$ is the predicted probability that the true coefficient $\beta_i$ corresponding to $i$-th covariate equals $\nu$. Then, we are in a binary prediction setting where each $\tilde{\beta}_i$ predicts $\beta_i = \nu$ for all $i \in \{1,\ldots,d \}$. We use the resulting AUC to assess the variable selection obtained through $\hat{\beta}$.

\section{Extended simulation results \label{sec:simulation results}}
Table~\ref{table:C-index simu appendix} bellow presents the results of simulation for the configurations $(d,r_c)=(30,0.2)$, (100,0.2) and (100,0.5).
\begin{sidewaystable}[!htbp]
\centering
\captionsetup{justification=justified}
\caption{Average C-index and standard deviation (in parenthesis) on 100 simulated data for different configurations $(d,r_c)$, with geometric distributions for the C-mix model. For each configuration, the best result appears in bold.}
\resizebox{\textwidth}{!}{%
\begin{tabular}{ccccccccccc}
\multicolumn{11}{c@{}}{$(d,r_c)=(30,0.2)$}\\
\midrule
& & \multicolumn{9}{c@{}}{$\textnormal{Estimation}$}\\
\cmidrule(l){3-11}
& & \multicolumn{3}{c@{}}{ \hspace{-.7cm} $n=100$ \hspace{.7cm}} & \multicolumn{3}{c@{}}{\hspace{-1.2cm} $n=200$ \hspace{.7cm}} & \multicolumn{3}{c@{}}{$n=500$} \\
Simulation & gap \hspace{.8cm} & C-mix & CURE & \hspace{-.5cm} Cox\ PH \hspace{.5cm} & C-mix & CURE & \hspace{-.5cm} Cox\ PH \hspace{.5cm} & C-mix & CURE & Cox\ PH \\
\midrule
 & 0.1 \hspace{.8cm} & \textbf{0.753 (0.055)} & 0.637 (0.069) & 0.658 (0.081) \hspace{.8cm} & \textbf{0.762 (0.034)} & 0.664 (0.070) & 0.704 (0.051) \hspace{.8cm} & \textbf{0.767 (0.023)} & 0.686 (0.062) & 0.749 (0.025) \\
C-mix & 0.3 \hspace{.8cm} & \textbf{0.756 (0.050)} & 0.599 (0.073) & 0.657 (0.075) \hspace{.8cm} & \textbf{0.761 (0.033)} & 0.600 (0.064) & 0.713 (0.050) \hspace{.8cm} & \textbf{0.757 (0.020)} & 0.565 (0.049) & 0.740 (0.021) \\
 & 1 \hspace{.8cm} & \textbf{0.723 (0.059)} & 0.710 (0.063) & 0.714 (0.062) \hspace{.8cm} & \textbf{0.723 (0.042)} & 0.718 (0.044) & 0.721 (0.040) \hspace{.8cm} & \textbf{0.727 (0.026)} & 0.723 (0.028) & 0.726 (0.025) \\
 \midrule
  & 0.1 \hspace{.8cm} & \textbf{0.918 (0.042)} & 0.872 (0.070) & 0.850 (0.081) \hspace{.8cm} & \textbf{0.938 (0.022)} & 0.911 (0.032) & 0.906 (0.034) \hspace{.8cm} & \textbf{0.949 (0.014)} & 0.940 (0.018) & 0.938 (0.017) \\
Cox PH & 0.3 \hspace{.8cm} & \textbf{0.935 (0.034)} & 0.906 (0.051) & 0.877 (0.066) \hspace{.8cm} & \textbf{0.947 (0.019)} & 0.932 (0.028) & 0.915 (0.030) \hspace{.8cm} & \textbf{0.952 (0.013)} & 0.950 (0.015) & 0.949 (0.015) \\
 & 1 \hspace{.8cm} & 0.956 (0.031) & \textbf{0.958 (0.032)} & 0.919 (0.065) \hspace{.8cm} & 0.960 (0.018) & \textbf{0.969 (0.016)} & 0.951 (0.024) \hspace{.8cm} & 0.958 (0.011) & \textbf{0.968 (0.011)} & 0.967 (0.010) \\
\bottomrule
\\
\multicolumn{11}{c@{}}{$(d,r_c)=(100,0.2)$}\\
\midrule
& & \multicolumn{9}{c@{}}{$\textnormal{Estimation}$}\\
\cmidrule(l){3-11}
& & \multicolumn{3}{c@{}}{ \hspace{-.7cm} $n=100$ \hspace{.7cm}} & \multicolumn{3}{c@{}}{\hspace{-1.2cm} $n=200$ \hspace{.7cm}} & \multicolumn{3}{c@{}}{$n=500$} \\
Simulation & gap \hspace{.8cm} & C-mix & CURE & \hspace{-.5cm} Cox\ PH \hspace{.5cm} & C-mix & CURE & \hspace{-.5cm} Cox\ PH \hspace{.5cm} & C-mix & CURE & Cox\ PH \\
\midrule
 & 0.1 \hspace{.8cm} & \textbf{0.736 (0.048)} & 0.601 (0.081) & 0.656 (0.066) \hspace{.8cm} & \textbf{0.757 (0.037)} & 0.629 (0.079) & 0.697 (0.057) \hspace{.8cm} & \textbf{0.767 (0.020)} & 0.659 (0.073) & 0.744 (0.024) \\
C-mix & 0.3 \hspace{.8cm} & \textbf{0.733 (0.056)} & 0.582 (0.063) & 0.648 (0.073) \hspace{.8cm} & \textbf{0.757 (0.035)} & 0.572 (0.047) & 0.699 (0.057) \hspace{.8cm} & \textbf{0.758 (0.023)} & 0.558 (0.040) & 0.736 (0.031) \\
 & 1 \hspace{.8cm} & \textbf{0.723 (0.067)} & 0.717 (0.073) & 0.705 (0.063) \hspace{.8cm} & \textbf{0.721 (0.041)} & 0.716 (0.041) & 0.719 (0.046) \hspace{.8cm} & 0.724 (0.023) & 0.720 (0.025) & \textbf{0.726 (0.023)} \\
 \midrule
  & 0.1 \hspace{.8cm} & \textbf{0.892 (0.047)} & 0.818 (0.086) & 0.830 (0.085) \hspace{.8cm} & \textbf{0.935 (0.026)} & 0.896 (0.048) & 0.904 (0.041) \hspace{.8cm} & \textbf{0.948 (0.013)} & 0.935 (0.021) & 0.940 (0.015) \\
Cox PH & 0.3 \hspace{.8cm} & \textbf{0.914 (0.042)} & 0.858 (0.076) & 0.869 (0.077) \hspace{.8cm} & \textbf{0.937 (0.025)} & 0.909 (0.038) & 0.917 (0.030) \hspace{.8cm} & \textbf{0.957 (0.011)} & 0.951 (0.014) & 0.951 (0.012) \\
 & 1 \hspace{.8cm} & 0.921 (0.040) & \textbf{0.937 (0.036)} & 0.917 (0.045) \hspace{.8cm} & 0.918 (0.033) & 0.947 (0.035) & \textbf{0.951 (0.024)} \hspace{.8cm} & 0.915 (0.018) & 0.959 (0.022) & \textbf{0.964 (0.011)} \\
\bottomrule
\\
\multicolumn{11}{c@{}}{$(d,r_c)=(100,0.5)$}\\
\midrule
& & \multicolumn{9}{c@{}}{$\textnormal{Estimation}$}\\
\cmidrule(l){3-11}
& & \multicolumn{3}{c@{}}{ \hspace{-.7cm} $n=100$ \hspace{.7cm}} & \multicolumn{3}{c@{}}{\hspace{-1.2cm} $n=200$ \hspace{.7cm}} & \multicolumn{3}{c@{}}{$n=500$} \\
Simulation & gap \hspace{.8cm} & C-mix & CURE & \hspace{-.5cm} Cox\ PH \hspace{.5cm} & C-mix & CURE & \hspace{-.5cm} Cox\ PH \hspace{.5cm} & C-mix & CURE & Cox\ PH \\
\midrule
 & 0.1 \hspace{.8cm} & \textbf{0.773 (0.064)} & 0.710 (0.087) & 0.678 (0.078) \hspace{.8cm} & \textbf{0.798 (0.038)} & 0.767 (0.057) & 0.744 (0.055) \hspace{.8cm} & \textbf{0.804 (0.022)} & 0.795 (0.024) & 0.788 (0.025) \\
C-mix & 0.3 \hspace{.8cm} & \textbf{0.781 (0.057)} & 0.696 (0.103) & 0.697 (0.087) \hspace{.8cm} & \textbf{0.798 (0.034)} & 0.741 (0.064) & 0.741 (0.055) \hspace{.8cm} & \textbf{0.800 (0.021)} & 0.778 (0.036) & 0.785 (0.023) \\
 & 1 \hspace{.8cm} & \textbf{0.772 (0.064)} & 0.742 (0.081) & 0.760 (0.071) \hspace{.8cm} & \textbf{0.772 (0.044)} & 0.732 (0.074) & 0.771 (0.041) \hspace{.8cm} & 0.770 (0.028) & 0.740 (0.059) & \textbf{0.771 (0.029)} \\
 \midrule
 & 0.1 \hspace{.8cm} & 0.755 (0.070) & \textbf{0.759 (0.068)} & 0.692 (0.082) \hspace{.8cm} & 0.780 (0.044) & \textbf{0.782 (0.043)} & 0.752 (0.052) \hspace{.8cm} & \textbf{0.795 (0.025)} & \textbf{0.795 (0.025)} & 0.785 (0.026) \\
CURE & 0.3 \hspace{.8cm} & 0.730 (0.077) & \textbf{0.737 (0.076)} & 0.674 (0.086) \hspace{.8cm} & 0.740 (0.042) & \textbf{0.740 (0.041)} & 0.708 (0.055) \hspace{.8cm} & 0.753 (0.028) & \textbf{0.753 (0.027)} & 0.740 (0.031) \\
 & 1 \hspace{.8cm} & \textbf{0.663 (0.075)} & 0.660 (0.076) & 0.659 (0.064) \hspace{.8cm} & 0.661 (0.053) & \textbf{0.661 (0.052)} & 0.658 (0.050) \hspace{.8cm} & \textbf{0.657 (0.032)} & 0.657 (0.033) & 0.657 (0.034) \\
 \midrule
  & 0.1 \hspace{.8cm} & 0.916 (0.069) & \textbf{0.924 (0.056)} & 0.837 (0.097) \hspace{.8cm} & \textbf{0.950 (0.028)} & 0.949 (0.029) & 0.911 (0.052) \hspace{.8cm} & \textbf{0.964 (0.012)} & \textbf{0.964 (0.012)} & 0.951 (0.016) \\
Cox PH & 0.3 \hspace{.8cm} & \textbf{0.937 (0.047)} & 0.934 (0.050) & 0.863 (0.071) \hspace{.8cm} & 0.955 (0.026) & \textbf{0.956 (0.022)} & 0.925 (0.037) \hspace{.8cm} & \textbf{0.968 (0.012)} & \textbf{0.968 (0.012)} & 0.958 (0.015) \\
 & 1 \hspace{.8cm} & 0.963 (0.029) & \textbf{0.967 (0.027)} & 0.973 (0.024) \hspace{.8cm} & 0.966 (0.019) & 0.970 (0.017) & \textbf{0.984 (0.012)} \hspace{.8cm} & 0.962 (0.012) & 0.966 (0.011) & \textbf{0.988 (0.006)} \\
\bottomrule
\end{tabular}}
\label{table:C-index simu appendix}
\end{sidewaystable}

\section{Selected genes per model on the TCGA datasets\label{sec:selected genes}}

In Tables~\ref{table:selected genes BRCA},~\ref{table:selected genes GBM} and ~\ref{table:selected genes KIRC} hereafter, we detail the 20 most significant covariates for each model and for the three considered datasets. For each selected gene, we precise the corresponding effect in percentage, where we define the effect of covariate $j$ as $100 \times |\beta_j|\ /\ \norm{\beta}_1\ \%$.
Then, to explore physiopathological and epidemiological background that could explain the role of the selected genes in cancer prognosis, we search in MEDLINE (search performed on the 15th september 2016 at \href{http://www.nlm.nih.gov/bsd/pmresources.html}%
{http://www.nlm.nih.gov/bsd/pmresources.html}) the number of publications for different requests: $(1)$ selected gene name (e.g. UBTF), $(2)$ selected gene name and cancer (e.g. UBTF AND cancer[MesH]), $(3)$ selected gene name and cancer survival (e.g. UBTF AND cancer[MesH] AND survival).
We then estimate f$_1$ defined here as the frequency of publication dealing with cancer among all publications for this gene, $i.e.\ (2)/(1)$, and f$_2$ defined as the frequency of publication dealing with survival among publications dealing with cancer, $i.e.\ (3)/(2)$. 
A f$_1$ (respectively f$_2$) close to 1 just informs that the corresponding gene is well known to be highly related to cancer (respectively to cancer survival) by the genetic research community. 
Note that the CURE and Cox PH models tend to have a smaller support than the C-mix one, since they tend to select less than 20 genes. 

\begin{table}[!htbp]
\centering
\captionsetup{justification=justified}
\caption{Top 20 selected genes per model for the BRCA cancer, with the corresponding effects. Dots ($\cdot$) mean zeros.}
\begin{tabular}{cccccccccccc}
\toprule
Genes & \multicolumn{3}{c@{}}{Model effects (\%)} & \hspace{.3cm} & \multicolumn{3}{c@{}}{MEDLINE data}  \\
\cmidrule(l){2-4} \cmidrule(l){6-8}
& $\textnormal{C-mix}$ & $\textnormal{CURE}$ & $\textnormal{Cox PH}$ & \hspace{.3cm} & $(1)$ & f$_1$ & f$_2$\\
\midrule
PHKB$|$5257 & 9.8 & 7.2 & 4.3 & \hspace{.3cm} & 1079 & 0.20 & 0.37 \\
UBTF$|$7343 & 7.8 & 5.8 & 21.7 & \hspace{.3cm} & 14 & 0,21 & $\cdot$ \\
LOC100132707 & 5.7 & 3.9 & 18.8 & \hspace{.3cm} & $\cdot$ & $\cdot$ & $\cdot$ \\
CHTF8$|$54921 & 4.4 & $\cdot$ & 7.2 & \hspace{.3cm} & 1 & 1 & $\cdot$ \\
NFKBIA$|$4792 & 4.3 & 1.9 & 3.4 & \hspace{.3cm} & 247 & 0.27 & 0.22 \\
EPB41L4B$|$54566 & 3.6 & 2.6 & $\cdot$ & \hspace{.3cm} & 19 & 0.47 & 0.22 \\
UGP2$|$7360 & 3.6 & 2.2 & $\cdot$ & \hspace{.3cm} & 19 & 0.15 & 1 \\
DPY19L2P1$|$554236 & 3.3 & $\cdot$ & 3.3 & \hspace{.3cm} & 1 & $\cdot$ & $\cdot$ \\
TRMT2B$|$79979 & 3.3 & 2.2 & $\cdot$ & \hspace{.3cm} & $\cdot$ & $\cdot$ & $\cdot$ \\
HSD3B7$|$80270 & 3.2 & 1.9 & 7.6 & \hspace{.3cm} & 19 & 0.05 & $\cdot$ \\
DLAT$|$1737 & 3.2 & 2.9 & $\cdot$ & \hspace{.3cm} & 75 & 0.16 & 0.16 \\
NIPAL2$|$79815 & 2.8 & 1.9 & $\cdot$ & \hspace{.3cm} & $\cdot$ & $\cdot$ & $\cdot$ \\
FGD3$|$89846 & 2.7 & $\cdot$ & 5.9 & \hspace{.3cm} & 10 & 0.2 & 0.5 \\
JRKL$|$8690 & 2.7 & 2.6 & $\cdot$ & \hspace{.3cm} & 2 & $\cdot$ & $\cdot$ \\
ZBED1$|$9189 & 2.5 & 2.4 & $\cdot$ & \hspace{.3cm} & 6 & $\cdot$ & $\cdot$ \\
KCNJ11$|$3767 & 2.3 & $\cdot$ & $\cdot$ & \hspace{.3cm} & 647 & 0.02 & $\cdot$ \\
WAC$|$51322 & 2.0 & 3.2 & $\cdot$ & \hspace{.3cm} & 260 & 0.05 & 0.25 \\
FLT3$|$2322 & 2.0 & $\cdot$ & $\cdot$ & \hspace{.3cm} & 4435 & 0.55 & 0.42 \\
STK3$|$6788 &  1.9 & 2.3 & $\cdot$ & \hspace{.3cm} & 107 & 0.32 & 0.15 \\
PAOX$|$196743 & 1.9 & 1.9 & $\cdot$ & \hspace{.3cm} & 18 & 0.11 & $\cdot$ \\
C14orf68$|$283600 & $\cdot$ & 3.3 & $\cdot$ & \hspace{.3cm} & $\cdot$ & $\cdot$ & $\cdot$ \\
LIN7C$|$55327 & $\cdot$ & 3.1 & $\cdot$ & \hspace{.3cm} & 36 & 0.06 & $\cdot$ \\
PNRC2$|$55629 & $\cdot$ & 2.1 & $\cdot$ & \hspace{.3cm} & 15 & $\cdot$ & $\cdot$ \\
SLC39A7$|$7922 & $\cdot$ & 1.8 & $\cdot$ & \hspace{.3cm} & 22 & 0.18 & $\cdot$ \\
MAGT1$|$84061 & $\cdot$ & 1.7 & $\cdot$ & \hspace{.3cm} & 50 & 0.12 & 0.17 \\
IRF2$|$3660 & $\cdot$ & $\cdot$ & 10.9 & \hspace{.3cm} & 310 & 0.21 & 0.14 \\
PELO$|$53918 & $\cdot$ & $\cdot$ & 7.0 & \hspace{.3cm} & 265 & 0.08 & 0.04 \\
SUSD3$|$203328 & $\cdot$ & $\cdot$ & 5.3 & \hspace{.3cm} & 5 & 0.6 & 0.67 \\
LEF1$|$51176 & $\cdot$ & $\cdot$ & 3.2 & \hspace{.3cm} & 940 & 0.29 & 0.23 \\
CPA4$|$51200 & $\cdot$ & $\cdot$ & 1.4 & \hspace{.3cm} & 18 & 0.22 & $\cdot$ \\
\bottomrule
\end{tabular}
\label{table:selected genes BRCA}
\end{table}

\begin{table}[!htbp]
\centering
\captionsetup{justification=justified}
\caption{Top 20 selected genes per model for the GBM cancer, with the corresponding effects. Dots ($\cdot$) mean zeros.}
\begin{tabular}{cccccccccccc}
\toprule
Genes & \multicolumn{3}{c@{}}{Model effects (\%)} & \hspace{.3cm} & \multicolumn{3}{c@{}}{MEDLINE data}  \\
\cmidrule(l){2-4} \cmidrule(l){6-8}
& $\textnormal{C-mix}$ & $\textnormal{CURE}$ & $\textnormal{Cox PH}$ & \hspace{.3cm} & $(1)$ & f$_1$ & f$_2$\\
\midrule
ARMCX6$|$54470 & 4.9 & $\cdot$ & 23.6 & \hspace{.3cm} & 1 & $\cdot$ & $\cdot$ \\
FAM35A$|$54537 & 4.4 & $\cdot$ & 21.8 & \hspace{.3cm} & $\cdot$ & $\cdot$ & $\cdot$ \\
CLEC4GP1$|$440508 & 3.9 & 5.1 & 2.8 & \hspace{.3cm} & $\cdot$ & $\cdot$ & $\cdot$ \\
INSL3$|$3640 & 3.6 & 2.7 & 1.7 & \hspace{.3cm} & 404 & 0.06 & 0.12 \\
REM1$|$28954 & 3.2 & $\cdot$ & $\cdot$ & \hspace{.3cm} & 54 & 0.05 & 0.66 \\
FAM35B2$|$439965 & 3.0 & $\cdot$ & $\cdot$ & \hspace{.3cm} & $\cdot$ & $\cdot$ & $\cdot$ \\
TSPAN4$|$7106 & 2.7 & $\cdot$ & $\cdot$ & \hspace{.3cm} & 16 & 0.31 & 0.4 \\
AP3M1$|$26985 & 2.7 & $\cdot$ & $\cdot$ & \hspace{.3cm} & 2 & 0.5 & $\cdot$ \\
PXN$|$5829 & 2.6 & $\cdot$ & 15.4 & \hspace{.3cm} & 891 & 0.25 & 0.18 \\
PDE4C$|$5143 & 2.5 & $\cdot$ & $\cdot$ & \hspace{.3cm} & 67 & 0.06 & 0.25 \\
PGBD5$|$79605 &  2.5 & $\cdot$ & $\cdot$ & \hspace{.3cm} & 5 & 0.25 & $\cdot$ \\
NRG1$|$3084 & 2.4 & $\cdot$ & 18.5 & \hspace{.3cm} & 1207 & 0.12 & 0.29 \\
LOC653786 & 2.2 & $\cdot$ & $\cdot$ & \hspace{.3cm} & $\cdot$ & $\cdot$ & $\cdot$ \\
FERMT1$|$55612 & 2.1 & $\cdot$ & $\cdot$ & \hspace{.3cm} & 115 & 0.19 & 0.18 \\
PLD3$|$23646 & 2.0 & $\cdot$ & $\cdot$ & \hspace{.3cm} & 38 & 0.10 & 0.25 \\
MIER1$|$57708 & 1.9 & $\cdot$ & 2.1 & \hspace{.3cm} & 16 & 0.31 & $\cdot$ \\
UTP14C$|$9724 & 1.8 & $\cdot$ & $\cdot$ & \hspace{.3cm} & 5 & 0.4 & $\cdot$ \\
AZU1$|$566 & 1.8 & $\cdot$ & $\cdot$ & \hspace{.3cm} & 15 & 0.2 & 0.33 \\
KCNC4$|$3749 & 1.7 & $\cdot$ & $\cdot$ & \hspace{.3cm} & 30 & 0.1 & 0.33 \\
FAM35B$|$414241 & 1.6 & $\cdot$ & $\cdot$ & \hspace{.3cm} & $\cdot$ & $\cdot$ & $\cdot$ \\
CRELD1$|$78987 & $\cdot$ & 32.2 & $\cdot$ & \hspace{.3cm} & 32 & 0.03 & $\cdot$ \\
HMGN5$|$79366 & $\cdot$ & 21.2 & $\cdot$ & \hspace{.3cm} & 41 & 0.54 & 0.32 \\
PNLDC1$|$154197 & $\cdot$ & 12.2 & $\cdot$ & \hspace{.3cm} & 3 & $\cdot$ & $\cdot$ \\
LOC493754 & $\cdot$ & 9.8 & $\cdot$ & \hspace{.3cm} & $\cdot$ & $\cdot$ & $\cdot$ \\
KIAA0146$|$23514 & $\cdot$ & 8.7 & $\cdot$ & \hspace{.3cm} & 3 & 0.67 & $\cdot$ \\
TMCO6$—$55374 & $\cdot$ & 3.6 & $\cdot$ & \hspace{.3cm} & 4 & 0.25 & $\cdot$ \\
ABLIM1$|$3983 & $\cdot$ & 2.1 & $\cdot$ & \hspace{.3cm} & 20 & 0.2 & $\cdot$ \\
OSBPL11$|$114885 & $\cdot$ & 1.0 & $\cdot$ & \hspace{.3cm} & $\cdot$ & $\cdot$ & $\cdot$ \\
TRAPPC1$|$58485 & $\cdot$ & 0.9 & $\cdot$ & \hspace{.3cm} & 4 & 0.75 & $\cdot$ \\
TBCEL$|$219899 & $\cdot$ & 0.5 & $\cdot$ & \hspace{.3cm} & 7 & 0.28 & $\cdot$ \\
RPL39L$|$116832 & $\cdot$ & $\cdot$ & 8.8 & \hspace{.3cm} & 10 & 0.7 & 0.14 \\
GALE$|$2582 & $\cdot$ & $\cdot$ & 3.5 & \hspace{.3cm} & 540 & 0.02 & $\cdot$ \\
BBC3$|$27113 & $\cdot$ & $\cdot$ & 0.7 & \hspace{.3cm} & 561 & 0.54 & 0.38 \\
DUSP6$|$1848 & $\cdot$ & $\cdot$ & 0.6 & \hspace{.3cm} & 307 & 0.30 & 0.22 \\
\bottomrule
\end{tabular}
\label{table:selected genes GBM}
\end{table}

\begin{table}[!htbp]
\centering
\captionsetup{justification=justified}
\caption{Top 20 selected genes per model for the KIRC cancer, with the corresponding effects. Dots ($\cdot$) mean zeros.}
\begin{tabular}{cccccccccccc}
\toprule
Genes & \multicolumn{3}{c@{}}{Model effects (\%)} & \hspace{.3cm} & \multicolumn{3}{c@{}}{MEDLINE data}  \\
\cmidrule(l){2-4} \cmidrule(l){6-8}
& $\textnormal{C-mix}$ & $\textnormal{CURE}$ & $\textnormal{Cox PH}$ & \hspace{.3cm} & $(1)$ & f$_1$ & f$_2$\\
\midrule
BCL2L12$|$83596 & 8.6 & 2.7 & $\cdot$ & \hspace{.3cm} & 64 & 0.72 & 0.39 \\
MARS$|$4141 & 7.5 & 6.9 & 7.2 & \hspace{.3cm} & 577 & 0.02 & 0.1 \\
NUMBL$|$9253 & 7.2 & 28.6 & 3.3 & \hspace{.3cm} & 56 & 0.14 & 0.25 \\
CKAP4$|$10970 & 6.1 & 10.6 & 22.3 & \hspace{.3cm} & 825 & 0.63 & 0.11 \\
HN1$|$51155 & 5.8 & 3.8 & $\cdot$ & \hspace{.3cm} & 13 & 0.38 & 0.2 \\
GIPC2$|$54810 & 5.7 & $\cdot$ & $\cdot$ & \hspace{.3cm} & 15 & 0.6 & 0.11 \\
NPR3$|$4883 & 5.2 & $\cdot$ & $\cdot$ & \hspace{.3cm} & 105 & 0.05 & 0.6 \\
GBA3$|$57733 &  5.0 & $\cdot$ & $\cdot$ & \hspace{.3cm} & 19 & 0.10 & $\cdot$ \\
SLC47A1$|$55244 & 5.0 & $\cdot$ & $\cdot$ & \hspace{.3cm} & 70 & 0.06 & $\cdot$ \\
ALDH3A2$|$224 & 4.7 & $\cdot$ & 2.6 & \hspace{.3cm} & 52 & 0.06 & 0.33 \\
CCNF$|$899 & 4.2 & 2.8 & $\cdot$ & \hspace{.3cm} & 50 & 0.24 & 0.08 \\
EHHADH$|$1962 & 3.9 & $\cdot$ & $\cdot$ & \hspace{.3cm} & 90 & 0.1 & $\cdot$ \\
SGCB$|$6443 & 3.3 & $\cdot$ & $\cdot$ & \hspace{.3cm} & 30 & $\cdot$ & $\cdot$ \\
GFPT2$|$9945 & 2.7 & 1.3 & $\cdot$ & \hspace{.3cm} & 18 & 0.22 & 0.25 \\
PPAP2B$|$8613 & 2.3 & $\cdot$ & $\cdot$ & \hspace{.3cm} & 29 & 0.17 & 0.2 \\
MBOAT7$|$79143 & 1.9 & 13.8 & 11.1 & \hspace{.3cm} & 15 & $\cdot$ & $\cdot$ \\
OSBPL1A$|$114876 & 1.5 & $\cdot$ & $\cdot$ & \hspace{.3cm} & 7 & $\cdot$ & $\cdot$ \\
C16orf57$|$79650 & 1.2 & $\cdot$ & $\cdot$ & \hspace{.3cm} & 26 & $\cdot$ & $\cdot$ \\
ATXN7L3$|$56970 & 0.9 & 2.5 & $\cdot$ & \hspace{.3cm} & 9 & $\cdot$ & $\cdot$ \\
C16orf59$|$80178 & 0.8 & $\cdot$ & $\cdot$ & \hspace{.3cm} & 3 & 0.66 & $\cdot$ \\
STRADA$—$92335 & $\cdot$ & 20.7 & 53.5 & \hspace{.3cm} & 9 & $\cdot$ & $\cdot$ \\
ABCC10$|$89845 & $\cdot$ & 3.9 & $\cdot$ & \hspace{.3cm} & 80 & 0.32 & 0.23 \\
MDK$|$4192 & $\cdot$ & 1.2 & $\cdot$ & \hspace{.3cm} & 789 & 0.38 & 0.23 \\
C16orf59$|$80178 & $\cdot$ & 1.1 & $\cdot$ & \hspace{.3cm} & 3 & 0.6 & $\cdot$ \\
\bottomrule
\end{tabular}
\label{table:selected genes KIRC}
\end{table}

\newpage

\bibliography{refs}
\bibliographystyle{plainnat}

\end{document}